\pdfoutput=1
\documentclass{article} %
\usepackage{iclr2027_conference,times}

\usepackage{amsmath,amsfonts,bm}

\def\eqref#1{equation~\ref{#1}}

\def\1{\bm{1}}

\DeclareMathAlphabet{\mathsfit}{\encodingdefault}{\sfdefault}{m}{sl}
\SetMathAlphabet{\mathsfit}{bold}{\encodingdefault}{\sfdefault}{bx}{n}

\usepackage{hyperref}
\usepackage[capitalize,nameinlink,noabbrev]{cleveref}  %
\usepackage{url}
\usepackage{xspace}
\usepackage{booktabs}
\usepackage{bbm}       %
\usepackage{palette}   %
\usepackage[most]{tcolorbox}   %
\newtcolorbox[auto counter, crefname={Takeaway}{Takeaways}]{takeawaybox}[1][]{colframe=petroil2,
  colback=lacamlilac!04!white, colbacktitle=petroil2, fonttitle=\bfseries,
  arc=0pt, outer arc=0pt, enhanced,
  attach boxed title to top left={yshift=-1pt},
  boxed title style={arc=0pt, outer arc=0pt}, #1}
\usepackage{tikz}
\usepackage{pgfplots}
\pgfplotsset{compat=1.18}
\usepgfplotslibrary{groupplots}
\usetikzlibrary{arrows.meta,positioning,calc,fit,backgrounds,decorations.pathreplacing,shapes.geometric}

\hypersetup{
    colorlinks=true,
    citecolor=teal,
    linkcolor=orange,
    urlcolor=teal,
    pdftitle={Merit: Scaling KG Foundation Models with Relation Transformers},
    pdfauthor={Cosimo Gregucci, Daniel Hern\'andez, Antonio Vergari, Steffen Staab},
}

\newcommand{\figfonts}{\renewcommand{\sfdefault}{cmss}\renewcommand{\bfdefault}{bx}\sffamily}

\tikzset{
  entity/.style     = {draw=#1, circle, line width=2pt, minimum size=14pt, inner sep=0pt},
  entity/.default   = bgrey1,
  dot/.style        = {fill=#1, circle, inner sep=0pt, minimum size=6pt},
  dot/.default      = bgrey2,
  boxnode/.style    = {draw=bgrey3, fill=bgrey3!12, rounded corners=3pt, inner sep=3pt,
                       align=center, font=\footnotesize\sffamily, text=bgrey4},
  badge/.style      = {draw=#1, fill=#1, rounded corners=1.5pt, inner sep=3pt,
                       font=\scriptsize\bfseries\sffamily, text=white},
  badge/.default    = tomato2,
  lbl/.style        = {text=bgrey4, font=\scriptsize\sffamily, inner sep=1pt},
  lblit/.style      = {text=bgrey4, font=\scriptsize\sffamily\itshape, inner sep=1pt},
  lbltiny/.style    = {text=bgrey4, font=\tiny\sffamily, inner sep=1pt},
  edge/.style       = {-{Stealth[length=5pt]}, bgrey2, line width=1.2pt},              %
  edgemiss/.style   = {densely dotted, -{Stealth[length=5pt]}, bgrey2, line width=1.2pt}, %
  edgepos/.style    = {-{Stealth[length=5pt]}, olive3, line width=1.6pt},              %
  edgeneg/.style    = {densely dotted, -{Stealth[length=5pt]}, tomato1, line width=1.6pt}, %
  edgethick/.style  = {line width=2.4pt},   %
  tipTri/.style     = {-{Triangle[length=3.4pt 1.1, width=3.4pt 1.1]}},
  tipSq/.style      = {-{Square[length=3.4pt 1.1, width=3.4pt 1.1]}},
  groupbox/.style   = {draw=gold4, dashed, rounded corners=4pt, inner sep=8pt},
  grouplbl/.style   = {font=\scriptsize\sffamily\itshape, text=gold6},
}

\newcommand{\Gi}{\mathcal{G}_\text{\textit{inf}}}               %
\newcommand{\Gu}{\mathcal{G}_t}               %
\newcommand{\Gtr}{\mathcal{G}_{tr}}  %
\newcommand{\Gri}{\mathcal{G}^r_\text{\textit{inf}}}            %

\newcommand{\Vi}{\mathcal{V}_\text{\textit{inf}}} 
\newcommand{\Vu}{\mathcal{V}_t}
\newcommand{\Rin}{\mathcal{R}_\text{\textit{inf}}} 
\newcommand{\Ru}{\mathcal{R}_t}
\newcommand{\Ei}{\mathcal{E}_\text{\textit{inf}}} 
\newcommand{\Eu}{\mathcal{E}_t}
\newcommand{\hth}{\text{h2h}}
\newcommand{\tth}{\text{t2h}}
\newcommand{\ttt}{\text{t2t}}
\newcommand{\htt}{\text{h2t}}

\newcommand{\MERIT}{MERIT\xspace}
\newcommand{\ULTRA}{ULTRA\xspace}
\newcommand{\MOTIF}{MOTIF\xspace}
\newcommand{\TRIX}{TRIX\xspace}
\newcommand{\Flock}{FLOCK\xspace}
\newcommand{\UltraQuery}{UltraQuery\xspace}
\newcommand{\MERITQuery}{MERITQuery\xspace}   %

\newcommand*{\fbnew}{{\footnotesize$\mathsf{FB15k237}$}\xspace}

\newcommand*{\fbnewH}{{\footnotesize$\mathsf{FB15k237}$+$\mathsf{H}$}\xspace}

\newcommand{\SQSA}{SQSA\xspace}
\newcommand{\SQUA}{SQUA\xspace}
\newcommand{\UQSA}{UQSA\xspace}
\newcommand{\UQUA}{UQUA\xspace}
\newcommand{\Orig}{Orig\xspace}
\newcommand{\halflink}{half-link\xspace}
\newcommand{\famone}{Family 1\xspace}
\newcommand{\famtwo}{Family 2\xspace}
\newcommand{\famthree}{Family 3\xspace}

\newcommand{\cg}[1]{\textbf{\textcolor{peas1}{{cg: #1}}}}

\usepackage{fontawesome5}
\definecolor{meritgold}{RGB}{212,175,55}
\DeclareRobustCommand{\meritmark}{{\color{meritgold}\faMedal}}
\newcommand\blfootnote[1]{%
  \begingroup
  \renewcommand\thefootnote{}\footnote{#1}%
  \addtocounter{footnote}{-1}%
  \endgroup
}

\newcommand{\resGpuMERIT}{10.73}

\newcommand{\stShareTimeMERITPre}{8.2}
\newcommand{\stShareTimeMERITRel}{4.8}

\newcommand{\stShareTimeTRIXRel}{84.5}

\newcommand{\stShareTimeMOTIFRel}{86.0}

\newcommand{\cqaTypesTicks}{1p,2p,3p,4p,2i,3i,4i,2i1p,1p2i,2u,2u1p,2in,3in,2in1p,2pi1pn,2nu1p}

\newcommand{\cqaEpUQFamOneAll}{0.062}
\newcommand{\cqaEpMQFamOneAll}{0.077}

\newcommand{\cqaEpUQFamTwoAll}{0.274}
\newcommand{\cqaEpMQFamTwoAll}{0.276}
\newcommand{\cqaEpUQFamTwoConj}{0.231}
\newcommand{\cqaEpMQFamTwoConj}{0.227}

\newcommand{\cqaEpUQFamTwoNeg}{0.316}
\newcommand{\cqaEpMQFamTwoNeg}{0.326}
\newcommand{\cqaEpUQFamThreeAll}{0.346}
\newcommand{\cqaEpMQFamThreeAll}{0.368}

\newcommand{\cqaEpUQAllAll}{0.278}
\newcommand{\cqaEpMQAllAll}{0.290}
\newcommand{\cqaEpUQAllConj}{0.263}
\newcommand{\cqaEpMQAllConj}{0.268}
\newcommand{\cqaEpUQAllUnion}{0.306}
\newcommand{\cqaEpMQAllUnion}{0.330}
\newcommand{\cqaEpUQAllNeg}{0.291}
\newcommand{\cqaEpMQAllNeg}{0.310}

\newcommand{\ablMrrUltra}{0.365}
\newcommand{\ablMrrMerit}{0.386}

\newcommand{\ablMrrPlain}{0.312}
\newcommand{\ablMrrMasked}{0.368}

\newcommand{\qphXmin}{1000}\newcommand{\qphXmax}{10000000}
\newcommand{\qphRatioUltraQueryTwoP}{1.94}
\newcommand{\qphRatioUltraQueryThreeP}{2.86}
\newcommand{\qphRatioMERITQueryTwoP}{1.96}
\newcommand{\qphRatioMERITQueryThreeP}{2.92}
\newcommand{\tcostAristoDeltaEight}{0.49}
\newcommand{\tcostAristoDeltaSixteen}{1.48}

\title{
Merit: Scaling KG Foundation Models \\with Relation Transformers
}

\author{Cosimo Gregucci\textsuperscript{1} \qquad Daniel Hernández\textsuperscript{1} \qquad Antonio Vergari\textsuperscript{2}\textsuperscript{\meritmark} \qquad Steffen Staab\textsuperscript{1}\textsuperscript{3}\textsuperscript{\meritmark}\\
\textsuperscript{1}Institute for Artificial Intelligence, University of Stuttgart, Germany\\
\textsuperscript{2}School of Informatics, University of Edinburgh, UK \qquad \textsuperscript{3}University of Southampton\\
\texttt{\{cosimo.gregucci,daniel.hernandez\}@ki.uni-stuttgart.de} \\
\texttt{\{steffen.staab\}@ki.uni-stuttgart.de} \\
\texttt{\{avergari\}@ed.ac.uk}
}

\iclrfinalcopy %
\begin{document}

\maketitle
\lhead{Preprint}

\begin{abstract}
Knowledge graph (KG) foundation models (KGFMs) predict links on unseen graphs zero-shot.
KGFMs have been pioneered by
ULTRA which builds a relation graph from co-occurrences that match motif templates, and passes messages over it to obtain relation representations.
More recent KGFMs such as MOTIF and TRIX increase the size of this relation graph, and FLOCK replaces message-passing with probabilistic random-walk ensembles.
However, larger relation graphs mean more messages, and more walks mean more sequences to encode, so they improve performance at higher training and inference cost.
This cost hinders their use in more complex reasoning tasks, such as complex query answering.
To reverse this trend, we design MERIT, the first KGFM with a transformer relation encoder: every relation name attends to every other, and the motifs enter as a soft prior rather than as edges.
Over $56$ benchmarks, MERIT matches the zero-shot MRR of SoTA KGFMs while being $6\times$ faster than TRIX and over $100\times$ faster than FLOCK at inference.
Fine-tuned for complex query answering like UltraQuery, MERITQuery improves over it 
at matched cost.
\blfootnote{\meritmark{} = shared supervision.}
\end{abstract}

\section{Introduction}
A knowledge graph (KG) stores facts as triples $(h,r,t)$, each linking a head entity $h$ to a tail entity $t$ through a relation name $r$. For instance, the fact that the Colosseo is a landmark of Rome is stored as $(\text{Rome}, \emph{hasLandmark}, \text{Colosseo})$ (\cref{fig:rel-graph}).
KGs support applications such as virtual assistants and recommendation~\citep{DBLP:conf/sigmod/IlyasRKPQS22,DBLP:conf/kdd/Dong18}, and are used to store incomplete data~\citep{DBLP:journals/csur/HoganBCdMGKGNNN21}.
To reason over incomplete KGs, link prediction answers a query $(h,r,?x)$ by ranking all candidate entities by the predicted plausibility that the resulting triple holds (\cref{sec:kgfms}).
KG foundation models (KGFMs) make link prediction zero-shot: trained once on a collection of source graphs, they predict links on unseen graphs without retraining~\citep{DBLP:conf/iclr/0001YM0Z24,DBLP:conf/icml/LeeCW23}.
In the \textit{inductive transfer setting}, a KGFM is conditioned at test time on a new inference graph $\Gi$, whose entities and relation names did not appear during pre-training, with edge set $\Ei$. Then, it answers queries $(h,r,?x)$ over a held-out edge set $\Eu$ that is disjoint from $\Ei$, but involving entities and relation names in $\Gi$, ranking every candidate entity $t$.
To answer queries zero-shot, entity and relation representations have to be computed regardless of specific entities or relation names. 

\ULTRA~\citep{DBLP:conf/iclr/0001YM0Z24} does so by building a relation graph $\Gri$ whose nodes correspond to relation names and whose directed edges are co-occurrences that match four hand-crafted motif templates (\cref{fig:rel-graph}), and by running message-passing on $\Gri$ with a relation encoder to obtain relation representations, on which it conditions a message-passing GNN entity encoder \citep{DBLP:conf/nips/ZhuZXT21} (\cref{sec:kgfms}).
The relation encoder passes messages only along the edges of $\Gri$, that is, only between relation names whose co-occurrence matches a hand-crafted motif template; we call this \textit{motif-constrained message-passing}.
\MOTIF~\citep{DBLP:conf/icml/HuangBBC0RO25} and \TRIX~\citep{DBLP:conf/log/ZhangBGR24} loosen it by increasing the size of the relation graph.
\Flock~\citep{DBLP:journals/corr/abs-2510-01510}, instead, avoids motif-constrained message-passing entirely, replacing the two message-passing steps with probabilistic random-walk ensembles. A larger relation graph means more messages, and in \Flock more walks mean more sequences to encode, so higher MRR than \ULTRA comes at higher training and inference costs.
Compared with \ULTRA, \textbf{\TRIX, \MOTIF and \Flock need $\boldsymbol{6\times}$ to $\boldsymbol{17\times}$ the pre-training time per batch and $\boldsymbol{6\times}$ to over $\boldsymbol{100\times}$ the inference time over the $\boldsymbol{56}$ benchmarks} (\cref{fig:1}).
This cost hinders their use in more complex reasoning tasks, such as complex query answering (CQA), where methods built on a neural link predictor run it at every hop of the query~\citep{DBLP:conf/iclr/ArakelyanDMC21,DBLP:conf/icml/BaiLLH23,DBLP:conf/iclr/Yin0S24} (\cref{sec:kgfms,app:qph}). We argue that this cost is why \UltraQuery~\citep{DBLP:conf/nips/GalkinZ00Z24}, which uses \ULTRA as its link predictor, is still the only KGFM for CQA, and there have not been equivalent versions for \TRIX, \MOTIF, or \Flock.
Behind these costs, the relation graphs of \ULTRA, \MOTIF and \TRIX are designed on the assumption that a set of hand-crafted motif templates covers the co-occurrences between relation names that matter for link prediction.
We argue that this assumption is too restrictive in two ways: two relation names that motif templates connect interact only through the templates that connect them, and two relation names that no template connects exchange no direct message.

Hence, we introduce \MERIT (\cref{sec:merit}), which replaces message-passing on the relation graph with a transformer over relation names, so that every relation name attends to every other and the motifs enter as a soft prior rather than as a constraint (\cref{fig:merit}).
How strongly two relation names interact is then learned from their representations and shifted by how often each motif occurs between them, so \MERIT keeps the information of the motifs that \ULTRA uses without being limited to it (\cref{sec:lp_results}).
Attention over all pairs grows quadratically with the number of relation names, but relation vocabularies are small, at most $3{,}210$ tokens with their inverses, on AristoV4 (\cref{tab:lp-stats}); fused kernels scale exact attention to $64$K tokens~\citep{DBLP:conf/nips/DaoFERR22}, also with an additive bias~\citep{DBLP:journals/corr/abs-2412-05496,DBLP:conf/nips/YingCLZKHSL21}.
\begin{figure}[t]
\centering
\input{figures/fig1v2}
\caption{\textbf{\MERIT matches SoTA KGFMs in MRR, with inference and pre-training times performance comparable to those of \ULTRA}, while \TRIX and \Flock need $6\times$ and over $100\times$
more inference time and $7\times$ and $6\times$ more training time,
respectively.
Each point is the mean MRR (higher is better) over $56$ benchmarks, averaged
over pre-training runs, five for \MERIT, \ULTRA and \TRIX and three for \MOTIF
and \Flock; bars are the sample standard deviation over runs. Panel (a) is
inference wall-clock over the whole suite and panel (b) pre-training time per
batch, both measured on an H100 GPU and shown on a log scale.}
\label{fig:1}
\end{figure}
Across $56$ benchmarks, \MERIT matches the zero-shot MRR of SoTA KGFMs with inference and pre-training times comparable to those of \ULTRA, the fastest existing KGFM, and with a similar number of FLOPs and peak GPU and CPU memory (\cref{fig:1,app:cost}).
Because \MERIT is as efficient as \ULTRA, it can be used for complex query answering.
\MERITQuery improves over \UltraQuery at matched cost (\cref{sec:cqa_results}). 
\paragraph{Contributions.} After reviewing how KGFMs reason over KGs (\cref{sec:kgfms}), 
(\textbf{C1}) we measure the training and inference cost of SoTA KGFMs, reporting FLOPs, peak GPU memory and inference time for the full suite of benchmarks, and pre-training time per batch (\cref{fig:1,app:cost}). 
(\textbf{C2}) We introduce \MERIT, a framework that allows all relation names to exchange messages while using the motifs as a soft prior (\cref{sec:merit}). 
(\textbf{C3}) We demonstrate that the inference time and memory of \MERIT stay comparable with those of \ULTRA at every graph size (\cref{sec:cost}), and that it matches SoTA KGFMs in zero-shot MRR at a fraction of their inference time (\cref{sec:lp_results}). 
(\textbf{C4}) We introduce new hardness-balanced benchmarks for CQA following \citet{DBLP:conf/icml/GregucciX0LMSV25} (\cref{app:plush}), and \MERITQuery, which improves over \UltraQuery at matched inference time (\cref{sec:cqa_results}).

\section{Reasoning over KGs with KGFMs}\label{sec:kgfms}
\paragraph{Knowledge graphs and reasoning over KGs.}
Formally, a KG is a tuple $\mathcal{G} = (\mathcal{V}, \mathcal{R}, \mathcal{E})$, where $\mathcal{V}$ is a finite set of entities, $\mathcal{R}$ is a finite set of relation names, and $\mathcal{E} \subseteq \mathcal{V} \times \mathcal{R} \times \mathcal{V}$ is a set of triples.
The main reasoning tasks over KGs studied in the literature are link prediction and complex query answering (CQA).
Given a head entity $h$ and a relation name $r$, \textbf{link prediction} ranks all candidate entities $e \in \mathcal{V}$ by the predicted plausibility that $(h,r,e)$ holds, so that the correct but unobserved target $t$ ranks as high as possible.
\textbf{CQA} extends link prediction to more complex reasoning tasks where the objective is to retrieve a set of answers to a logical query that poses conditions over entities and relation names in a KG.

\begin{figure}[t]
\renewcommand{\sfdefault}{cmss}
\renewcommand{\bfdefault}{bx}

\tikzset{
  vringV5/.style     = {draw=bgrey1, circle, inner sep=0pt, minimum size=10pt, line width=2pt},
  obsV5/.style       = {bgrey1, line width=1.0pt},
  relCompV5/.style   = {shorten >=0.5pt, -{Triangle[length=3.4pt 1.1, width=3.4pt 1.1]}},
  relHasV5/.style    = {shorten >=0.5pt, -{Square[length=3.4pt 1.1, width=3.4pt 1.1]}},
  relFoughtV5/.style = {shorten >=0.5pt, -{Diamond[length=4.2pt 1.1, width=3.4pt 1.1]}},
  qlabSmallV5/.style = {draw=tomato0!70!tomato1, fill=tomato0!70!tomato1, inner sep=3.5pt, rounded corners=1.5pt},
  rntriV5/.style = {regular polygon, regular polygon sides=3, draw=bgrey1,
                    fill=bgrey1, inner sep=1.2pt, minimum size=12pt},
  rnsqV5/.style  = {rectangle, draw=bgrey1, fill=bgrey1, minimum size=8pt},
  rncircV5/.style= {diamond, draw=bgrey1, fill=bgrey1, inner sep=0pt, minimum size=11pt},
  relmotif/.style= {-stealth, bgrey2, line width=1pt, shorten >=0.5pt},
  mlbl/.style    = {bgrey4, font=\tiny\sffamily\itshape, inner sep=1pt},
  rlbl/.style    = {bgrey4, font=\tiny\sffamily\itshape, inner sep=1pt},
  vlblV2/.style  = {bgrey4, font=\scriptsize\sffamily, inner sep=1pt},
}

\noindent
\begin{minipage}[c]{0.61\linewidth}
\centering
\scalebox{1.07}{\includegraphics{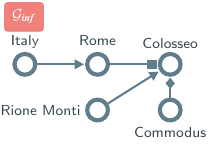}}\hspace{13pt}%
\scalebox{1.07}{\includegraphics{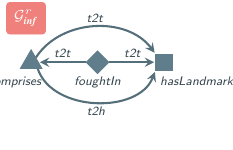}}\\
\scalebox{1.07}{\includegraphics{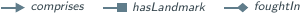}}
\end{minipage}\hfill
\begin{minipage}[c]{0.37\linewidth}
\caption{\textbf{Example inference graph $\Gi$ (left) and its ULTRA-derived
relation graph $\Gri$ (right).}
For visual simplification, we omit (i)~\textit{h2t} edges, since they
are the reverse of \textit{t2h}, and (ii)~trivial \textit{h2h} and
\textit{t2t} self-loops.}
\label{fig:rel-graph}
\label{fig:rel-graph:top}
\label{fig:rel-graph:bot}
\end{minipage}
\end{figure}

Following~\citet{DBLP:conf/iclr/ArakelyanDMC21,DBLP:conf/nips/RenL20}, we consider the problem of answering logical queries with a single \textit{target} variable, a set of constants including \textit{anchor} entities, given relation names, and first-order logical operations that include conjunction, disjunction, negation and existential quantification.
Complex queries are categorized into different types based on their structure;
the types we consider are explained and drawn in \cref{app:cqa-queries}, and the simplest of them is link prediction, which in this task is referred to as a \textit{1p} query.
A more complex query, a \textit{2i} query, involves an \textit{intersection} made of two links. Considering the graph $\Gi$ of \cref{fig:rel-graph}, and the query \textit{``Where did Commodus fight, among the places that Rione Monti comprises?''}, a link predictor would have to answer the two queries (Commodus, \emph{foughtIn}, $?x$), and (Rione Monti, \emph{comprises}, $?x$), and then combine the two answers. 

\paragraph{Reasoning on unseen graphs.}
Knowledge graph embeddings, such as TransE~\citep{DBLP:conf/nips/BordesUGWY13}, ComplEx~\citep{DBLP:conf/icml/TrouillonWRGB16} and RotatE~\citep{DBLP:conf/iclr/SunDNT19}, learn an embedding for every entity and relation name. GraIL~\citep{DBLP:conf/icml/TeruDH20} and NBFNet~\citep{DBLP:conf/nips/ZhuZXT21} generalize to unseen entities but still learn embeddings for relation names.
Hence, none of these methods can predict links in the \textit{inductive transfer setting}.
In this setting, a model is trained on multiple training graphs and, at test time, conditioned on a new inference graph $\Gi = (\Vi, \Rin, \Ei)$ whose entities and relation names do not appear during training~\citep{DBLP:conf/iclr/0001YM0Z24}.
The test graph $\Gu = (\Vu, \Ru, \Eu)$ holds the triples to predict, which are held out from $\Gi$ ($\Ei \cap \Eu = \emptyset$) but involve only entities and relation names of $\Gi$ ($\Vu \subseteq \Vi$, $\Ru \subseteq \Rin$).

\ULTRA
adds to $\Gi$ the inverse $(t, r^{-1}, h)$ of every triple $(h,r,t)$, for instance $(\text{Colosseo}, \emph{hasLandmark}^{-1}, \text{Rome})$, and builds a relation graph $\Gri$ in which every relation name in $\Rin$ and its inverse correspond each to a node. The inverses are omitted in \cref{fig:rel-graph} for visual simplification.
The edges of $\Gri$ are the co-occurrences that match the four motif templates \hth, \ttt, \htt{} and \tth, which connect two relation names when they share an entity as heads, as tails, or as the head of one and the tail of the other, in either direction.
For example, \emph{foughtIn} and \emph{comprises} share the tail Colosseo, so $\Gri$ links them by a \ttt{} edge (\cref{fig:rel-graph}).
Given a query $(h,p,?x)$ with query relation $p$, the relation encoder, a variant of NBFNet, runs motif-constrained message-passing on $\Gri$ to compute a representation of every relation name in $\Rin$, conditioned on $p$.
The entity encoder, another instance of NBFNet, runs message-passing on $\Gi$ with the relation representations as edge features; a final MLP then scores every candidate entity.

To loosen motif-constrained message-passing, \MOTIF~\citep{DBLP:conf/icml/HuangBBC0RO25} and \TRIX~\citep{DBLP:conf/log/ZhangBGR24} extend the relation graph of \ULTRA along orthogonal axes, the former through motif length, by building a relational hypergraph, the latter through per-entity granularity. \TRIX additionally couples the relation and entity encoders through iterative updates.
\Flock~\citep{DBLP:journals/corr/abs-2510-01510}, instead, removes message-passing entirely, and with it the motif templates: it iteratively samples random walks, encodes them into sequences, and embeds them with a sequence model, at a cost that grows with the number and the length of the walks.
However, these expressiveness gains come at a cost: all three require significantly more time for training and inference than \ULTRA (\cref{fig:1}).

\UltraQuery~\citep{DBLP:conf/nips/GalkinZ00Z24} adapts \ULTRA to CQA: it fine-tunes \ULTRA on complex queries and runs it at every hop, mapping its scores to a fuzzy set of entities with values in $[0,1]$ and combining these sets with fuzzy logic operators. See \cref{app:cqa-sota} for more related work on CQA.
In \cref{sec:merit}, we allow all relation names to exchange messages while using the motifs as a soft prior, not a constraint. We do so while preserving  the computational cost of \ULTRA, so that our \MERIT can scale to CQA tasks in the same way.

\section{\MERIT: Motif-Enriched Relational Inductive Transformer}\label{sec:merit}
\begin{figure}[t]
\noindent
\begin{minipage}[c]{0.59\linewidth}\centering
\scalebox{0.766}{%
{\figfonts
\tikzset{
  towerPL/.style={draw=#1!15, fill=#1!15, text=bgrey4,
    minimum width=2.7cm, minimum height=1.5cm, font=\normalsize\bfseries\sffamily,
    inner sep=6pt, align=center, line width=0.6pt},
  flowPL/.style={-{Stealth[length=5pt]}, bgrey2, line width=1pt},
  lblPL/.style={text=bgrey4, font=\small\sffamily},
  pics/freezePL/.style={code={
    \foreach \a in {0,60,120}{
      \draw[line width=0.8pt, rotate=\a] (0,-0.9) -- (0,0.9);
      \draw[line width=0.8pt, rotate=\a] (0,0.55) -- ++(30:0.3);
      \draw[line width=0.8pt, rotate=\a] (0,0.55) -- ++(150:0.3);
      \draw[line width=0.8pt, rotate=\a] (0,-0.55) -- ++(-30:0.3);
      \draw[line width=0.8pt, rotate=\a] (0,-0.55) -- ++(210:0.3);
    }
  }},
}
\includegraphics{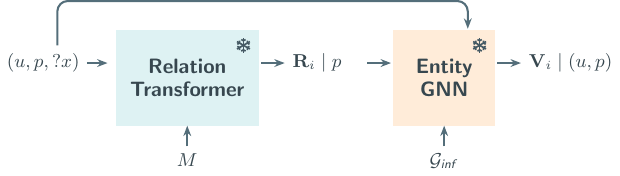}}}
\end{minipage}\hfill
\begin{minipage}[c]{0.38\linewidth}
\caption{\textbf{In \MERIT, the query-conditioned relation representations are obtained
with the relation transformer.} The rest of the pipeline is as in
\ULTRA (\cref{sec:kgfms}).}
\label{fig:pipeline}
\end{minipage}
\end{figure}
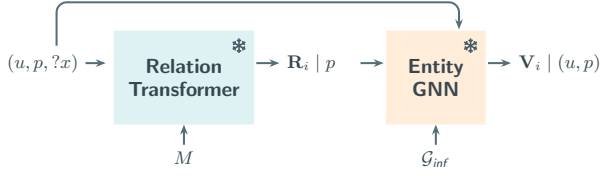

Under motif-constrained message-passing, two relation names exchange messages only if a motif template connects them (\cref{sec:kgfms}). The number of messages hence grows with the motifs, and with it the training and inference costs (\cref{fig:1}). In the following, we propose a framework where every relation name exchanges messages with every other, and where they can exchange messages that go beyond hand-crafted motifs.

\paragraph{\MERIT.} We introduce \MERIT (\textbf{M}otif-\textbf{E}nriched \textbf{R}elational \textbf{I}nductive \textbf{T}ransformer): a transformer over relation names in which the motifs enter as a soft prior rather than as the edges of a relation graph. The transformer of \MERIT reads $\Gi$ only through a \textbf{\textit{motif tensor}} $M \in \mathbb{R}^{|\Rin| \times |\Rin| \times T}$: the entry $M_{ij} \in \mathbb{R}^{T}$, the \textbf{\textit{pair descriptor}} of $(i,j)$, holds $T$ numbers that describe how relation names $i$ and $j$ co-occur in $\Gi$, and the row $M_i$ collects the descriptors of $i$ with the other relation names. The tensor enters at two points, the attention and the tokens,
\begin{equation}
M = \Psi(\Gi), \qquad m^{k}_{ij} = f_k(M_{ij}), \qquad \mathrm{s}_r = g\big(\{\!\{\, M_{rj} : j \in \Rin \setminus \{r\} \,\}\!\}\big) \in \mathbb{R}^{d}, \label{eq:framework}
\end{equation}
where $\Psi$ computes $M$ from $\Gi$, $f = (f_1, \ldots, f_K)\colon \mathbb{R}^{T} \to \mathbb{R}^{K}$ maps a pair descriptor to one \textbf{\textit{motif bias}} $m^k_{ij}$ for each of the $K$ heads, and $g$ maps the multiset of the pair descriptors of a relation name to a vector of $d$ entries, its \textbf{\textit{relation signature}} $\mathrm{s}_r$.
To ensure zero-shot generalization on unseen graphs, none of $\Psi$, $f$ and $g$ ties a parameter to a relation name. 
For every relation name $r$, the transformer outputs a representation $\mathbf{h}_{r \mid p}$ conditioned on the query relation $p$; an entity GNN, here NBFNet~\citep{DBLP:conf/nips/ZhuZXT21}, then computes the query-conditioned entity representations and scores for all entities as in \ULTRA~\citep{DBLP:conf/iclr/0001YM0Z24} (\cref{fig:pipeline}).

\begin{figure}[t]
\centering
{\figfonts
\tikzset{
  tokFM/.style     = {dot=petroil2, minimum size=9pt},
  btokFM/.style    = {dot=petroil2, minimum size=7pt},
  bframeFM/.style  = {rounded corners=4pt, bgrey0, line width=0.4pt},
  lthinFM/.style   = {bgrey0!45, line width=0.3pt, -{Latex[length=2.2pt, width=1.8pt]}, shorten >=1pt},
  lthickFM/.style  = {bgrey2, line width=0.8pt, -{Latex[length=3.2pt, width=2.6pt]}, shorten >=1pt},
  vecFM/.style     = {text=bgrey4, font=\scriptsize, inner sep=1pt, anchor=east},
  outFM/.style     = {text=bgrey4, font=\scriptsize, inner sep=1pt, anchor=west},
  idxFM/.style     = {text=bgrey4, font=\tiny\sffamily, inner sep=1pt},
  alblFM/.style    = {text=bgrey4, font=\fontsize{6}{7}\selectfont, inner sep=1pt},
  alblonFM/.style  = {text=bgrey4, font=\fontsize{6}{7}\selectfont, inner sep=0.3pt, fill=white, sloped},
  panelFM/.style   = {text=bgrey4, font=\scriptsize\sffamily, inner sep=1pt},
  rnodeFM/.style   = {text=bgrey4, font=\scriptsize, inner sep=1pt},
  rntriFM/.style   = {regular polygon, regular polygon sides=3, draw=bgrey1, fill=bgrey1, inner sep=1.0pt, minimum size=9pt},
  rnsqFM/.style    = {rectangle, draw=bgrey1, fill=bgrey1, minimum size=6.5pt},
  rndiaFM/.style   = {diamond, draw=bgrey1, fill=bgrey1, inner sep=0pt, minimum size=9pt},
  relmotifFM/.style= {-stealth, bgrey2, line width=0.8pt, shorten >=0.5pt},
  mlblFM/.style    = {bgrey4, font=\tiny\sffamily\itshape, inner sep=1pt},
  mhdrFM/.style    = {font=\tiny, text=bgrey4, inner sep=1pt},
  msqFM/.style     = {draw=bgrey0!60, fill=white, minimum size=0.20cm, inner sep=0pt, line width=0.3pt},
  msqonFM/.style   = {draw=bgrey0!60, fill=bgrey1, minimum size=0.20cm, inner sep=0pt, line width=0.3pt},
  msqsFM/.style    = {draw=bgrey0!60, fill=white, minimum size=0.10cm, inner sep=0pt, line width=0.3pt},
  msqsonFM/.style  = {draw=bgrey0!60, fill=bgrey1, minimum size=0.10cm, inner sep=0pt, line width=0.3pt},
}
\includegraphics{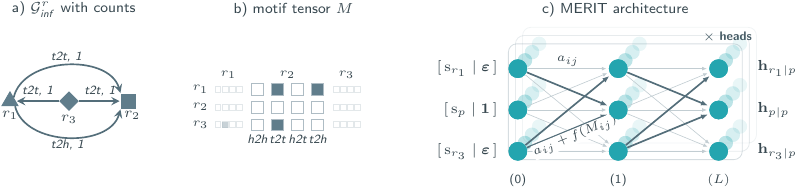}}
\caption{\textbf{\MERIT runs a transformer over all relation tokens and injects the motifs in the token embeddings and as a bias on the attention.} The motif tensor $M$ b) has the shape of the adjacency tensor of the relation graph $\Gri$ a), with entity occurrence counts in place of edges. This tensor is then used as a prior for the \MERIT architecture c), where every token contains the relation signature $\mathrm{s}_r$ of its relation, followed by random noise $\bm{\varepsilon}$, or by ones for the query relation $p$ ($r_2$ in the example); every arrow is an attention logit $a_{ij}$, and the arrows between co-occurring relations add the bias $f(M_{ij})$, over multiple heads. 
}
\label{fig:merit}
\end{figure}

\paragraph{Motif bias.} The transformer has $L$ layers, each with multi-head self-attention over all $|\Rin|$ tokens. The motifs enter here as additive bias \citep{DBLP:conf/nips/YingCLZKHSL21} on the attention logits: in each head $k$, the motif bias $m^k_{ij}$ of \cref{eq:framework} shifts the logit with which token $i$ attends to token $j$. We take $f(\mathbf{0}) = \mathbf{0}$, so that a pair between which no motif occurs, $M_{ij} = \mathbf{0}$, is attended on its query-key logit alone, as any pair in a plain transformer (thin arrows in \cref{fig:merit}c), while a pair between which motifs occur (thick arrows) is attended with a larger or a smaller weight. A further motif template adds a slice to $M$ and one entry to every pair descriptor, and leaves the token dimension, the heads, the layers and the attended pairs unchanged, so no connection between relation names is added. With $a^{(\ell,k)}_{ij}$ the query-key logit of head $k$ in layer $\ell$, the transformer reads
\begin{align}
\mathbf{x}^{(0)}_{r \mid p} &= \big[\, \mathrm{s}_r \;\big\|\; \mathbf{z}_{r \mid p} \,\big], \qquad \mathbf{z}_{r \mid p} = \mathbbm{1}_{r = p}\, \mathbf{1} + \mathbbm{1}_{r \neq p}\, \boldsymbol{\varepsilon}, \quad \boldsymbol{\varepsilon} \sim \mathcal{N}(\mathbf{0}, I), \label{eq:init}\\
\mathbf{x}^{(\ell+1)}_{i \mid p} &= \textsc{Update}\Big(\mathbf{x}^{(\ell)}_{i \mid p},\ \big\|_{k=1}^{K} \sum\nolimits_{j \in \Rin} \operatorname{softmax}_j\big(a^{(\ell,k)}_{ij} + m^k_{ij}\big)\, \textsc{Value}_k\big(\mathbf{x}^{(\ell)}_{j \mid p}\big)\Big), \label{eq:layer}
\end{align}
where $\|$ denotes concatenation, $\textsc{Value}_k$ is the value map of head $k$ and \textsc{Update} the output projection and residual feed-forward update of a pre-norm transformer encoder layer~\citep{DBLP:conf/nips/VaswaniSPUJGKP17,DBLP:conf/icml/XiongYHZZXZLWL20}, and $\mathbf{h}_{r \mid p} = \mathbf{x}^{(L)}_{r \mid p}$. The query flag $\mathbf{z}_{r \mid p}$ conditions the representations $\mathbf{h}_{r \mid p}$ on the query relation $p$: all ones on $p$ following the labelling trick of \ULTRA~\citep{DBLP:conf/iclr/0001YM0Z24}, and standard Gaussian noise on every other relation name, redrawn at every forward pass so that tokens with the same signature stay distinguishable~\citep{DBLP:conf/ijcai/AbboudCGL21}.

\paragraph{Relation signature.} The relation signature $\mathrm{s}_r$ describes relation name $r$ through its row of the motif tensor, independently of the query (\cref{fig:merit}c). Since $g$ reads the row as a multiset, $\mathrm{s}_r$ does not depend on the order of the other relation names and is defined for any number of them. In this way, two relation names that take part in the motifs in the same way receive similar tokens, and two that take part differently can receive different ones.

\paragraph{Instantiation.} For $\Psi$, we use the four motif templates \hth, \ttt, \htt{} and \tth{} of \cref{sec:kgfms}, so $T = 4$, and record in $M_{ijt}$ the compressed count of the occurrences of motif $t$ between $i$ and $j$, one per distinct entity. Our $f$ is linear, $f(M_{ij}) = B^\top M_{ij}$ with $B \in \mathbb{R}^{T \times K}$ shared by all layers, so motif $t$ raises the attention of head $k$ to a pair when $B_{tk} > 0$ and lowers it when $B_{tk} < 0$, by an amount that grows with the count; our $g$ is linear on a fixed summary of the row, $\mathrm{s}_r = W_g\,\phi(M_r)$, where $\phi(M_r) \in \mathbb{R}^{2T}$ holds, for each motif $t$, the largest entry $\max_{j \neq r} M_{rjt}$ of the row and the compressed count of the relation names $j$ with $M_{rjt} > 0$: how strongly and how broadly $r$ takes part in motif $t$. Both $B$ and $W_g$ start at zero, so at initialization \MERIT is a plain transformer over noise and the query flag, and the strength of the prior is learned. 
See \cref{app:impl} for more details.

\paragraph{\MERITQuery.} Mirroring \UltraQuery (\cref{sec:kgfms}), in \MERITQuery we use \MERIT as the link predictor over fuzzy sets of entities, with the logical connectives relaxed via fuzzy logic operators. \MERITQuery starts from pre-trained \MERIT and is fine-tuned on the complex queries of \fbnew, then answers queries on unseen KGs zero-shot (\cref{sec:cqa_results}).

\section{How does the cost of \MERIT scale?}\label{sec:cost}

\MERIT replaces message-passing on the relation graph with attention over all relation names, and further motifs add no connection between them (\cref{sec:merit}). We now measure the training and inference costs of \MERIT against those of SoTA KGFMs, and answer the following research question:
\textbf{(RQ1)} \textit{How do the training and inference costs of \MERIT scale, compared with those of SoTA KGFMs?} We study it along the size of the inference graph, the number of hops of a query and the number of motif templates.

\begin{figure}[t]
\noindent
\begin{minipage}[c]{0.58\linewidth}\centering
{\figfonts
\colorlet{mMERITsc}{olive5}
\colorlet{mULTRAsc}{petroil4}
\colorlet{mTRIXsc}{bdarker}
\colorlet{mMOTIFsc}{gold2}
\colorlet{mFlocksc}{lacamlilac}
\pgfplotsset{
  scalingaxisSC/.style={
    width=3.8cm, height=5.4cm,
    ymin=0.45, ymax=2200, ytick={1,10,100,1000},
    axis lines=left, axis line style={-, bgrey1, line width=0.5pt},
    tick style={bgrey1, line width=0.5pt}, grid=major,
    grid style={bgrey0!22, line width=0.4pt}, log ticks with fixed point,
    tick label style={font=\tiny\sffamily, text=bgrey4},
    label style={font=\scriptsize\sffamily, text=bgrey4},
    title style={font=\scriptsize\sffamily, text=bgrey4, yshift=-9pt},
    ylabel shift=-6pt,
    enlarge x limits=0.06, clip=false, unbounded coords=discard, xminorticks=false,
    every axis plot/.append style={line width=0.8pt},
  },
  sMERITSC/.style={mMERITsc, mark=pentagon*, mark size=2.0pt, mark options={fill=mMERITsc, draw=white, line width=0.4pt}},
  sULTRASC/.style={mULTRAsc, mark=*, mark size=1.6pt, mark options={fill=mULTRAsc, draw=white, line width=0.4pt}},
  sTRIXSC/.style={mTRIXsc, mark=triangle*, mark size=1.9pt, mark options={fill=mTRIXsc, draw=white, line width=0.4pt}},
  sMOTIFSC/.style={mMOTIFsc, mark=square*, mark size=1.5pt, mark options={fill=mMOTIFsc, draw=white, line width=0.4pt}},
  sFlockSC/.style={mFlocksc, mark=diamond*, mark size=1.9pt, mark options={fill=mFlocksc, draw=white, line width=0.4pt}},
}
\tikzset{
  lblSC/.style={font=\tiny\sffamily, inner sep=1pt},
  ourlblSC/.style={font=\tiny\bfseries\sffamily, inner sep=1pt, text=mMERITsc},
}
\includegraphics{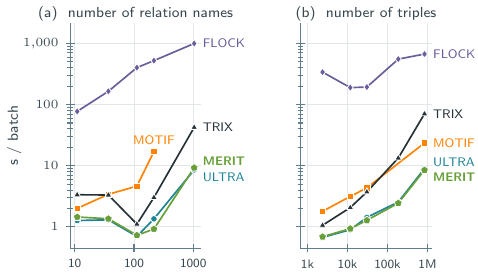}}
\end{minipage}\hfill
\begin{minipage}[c]{0.41\linewidth}
\caption{\textbf{\MERIT keeps the inference cost of \ULTRA at every graph
size}, while the cost gap to \TRIX widens with the number of triples and the gap
to \MOTIF with the number of relation names; \Flock stays two to three orders of magnitude above. Markers are
medians of the inference wall-clock within log-spaced bins of inference-graph size. Further analysis in \cref{fig:scaling-app},
bin details in \cref{tab:scaling-bins}. We omit the incomplete bins of \MOTIF from the plot.}
\label{fig:scaling}
\end{minipage}
\end{figure}

\paragraph{Experimental setup.}
We measure training and inference cost on one NVIDIA H100 GPU with $94$ GiB of memory, in a node with two AMD EPYC 9454 processors and $768$ GiB of CPU memory.
For the KGFMs with a relation encoder, all but \Flock, we also measure the resources of each stage of inference (\cref{app:stages}): the precomputation of the relational structure, $M$ for \MERIT; the relation encoder, for \MERIT the transformer; the entity encoder; and the score computation. \cref{fig:1,fig:stages} sum over the $56$ benchmarks other than AristoV4, the one benchmark on which we could not profile \MOTIF; \cref{fig:scaling} includes AristoV4 for every other model and omits the bins of \MOTIF that contain it. 
We measure pre-training time per batch to compare the speed of the models, not their time to convergence, which depends on the training recipe of each model.
\cref{app:batch} details the batch sizes and model configurations.

\paragraph{Costs of \MERIT stay comparable with those of \ULTRA at every graph size.}
Over the $56$ benchmarks, \MERIT needs $0.55$ hours of inference against $0.58$ for \ULTRA, $3.31$ for \TRIX, $4.20$ for \MOTIF and $61.86$ for \Flock, and $0.087$ seconds per pre-training batch against $0.089$, $0.584$, $1.518$ and $0.528$ (\cref{fig:1}).
Across bins of relation names, entities and triples of the inference graph, the inference time of \MERIT tracks that of \ULTRA (\cref{fig:scaling,app:scaling}).
The largest gap in seconds to \ULTRA is on AristoV4, $189$ against $153$ seconds, while \TRIX needs $605$ and \Flock $28{,}128$.
\Flock also needs $507$ GB of CPU memory on DBpedia100k, against at most $1.5$ GB for all others (\cref{fig:scaling,fig:scaling-app}).
The relation stage takes $\stShareTimeMERITRel\%$ of the inference time of \MERIT, against $\stShareTimeTRIXRel\%$ for \TRIX and $\stShareTimeMOTIFRel\%$ for \MOTIF (\cref{fig:stages}).
\paragraph{The inference time of \MERITQuery grows with the number of hops like that of \UltraQuery.}
On the path queries of \fbnew, a \textit{2p} query takes $\qphRatioMERITQueryTwoP$ times the time of a \textit{1p} query for \MERITQuery and $\qphRatioUltraQueryTwoP$ for \UltraQuery, and a \textit{3p} query $\qphRatioMERITQueryThreeP$ and $\qphRatioUltraQueryThreeP$ times, consistent with the link predictor running once per hop.
If a $k$-hop query takes $k$ times the time of a \textit{1p} query, \MERITQuery would answer more \textit{4p} queries in one hour than \TRIX, \MOTIF or \Flock answer \textit{1p} queries (\cref{fig:qph}).

\paragraph{Beyond precomputation, more motif templates leave the time of \MERIT unchanged.}
A further motif template adds to the precomputation of $M$, which takes $\stShareTimeMERITPre\%$ of the inference time of \MERIT (\cref{fig:stages}). It adds little to the relation stage and leaves the entity encoder unchanged.
To measure the relation stage at more templates, we duplicate the four slices of $M$ to $T = 8$ and $T = 16$: the pre-training time per batch and the inference time over all $57$ benchmarks, AristoV4 included, change by less than $1\%$ (\cref{app:tcost}).
On AristoV4, where $M$ is largest with $|\Rin|^2 T$ entries, peak GPU memory on the relation stage grows by only $\tcostAristoDeltaEight$ GB at $T = 8$ and $\tcostAristoDeltaSixteen$ GB at $T = 16$, and stays below the largest peak over the suite, $\resGpuMERIT$ GB, reached by the entity encoder.

\begin{takeawaybox}[label={takeaway:cost}]
\textbf{Takeaway 1.}
With a transformer over relation names, the cost of \MERIT stays comparable with that of \ULTRA at every graph size and number of hops of complex queries. Beyond precomputation, adding motif templates leaves the overall costs unchanged. We advise reporting training and inference cost alongside MRR when proposing new KGFMs.
\end{takeawaybox}

\section{How does \MERIT fare in KG reasoning?}\label{sec:results}
In this section, we evaluate \MERIT on link prediction and CQA, the two reasoning tasks of \cref{sec:kgfms}. 
In link prediction (\cref{sec:lp_results}), we ask whether \MERIT reaches the MRR of SoTA KGFMs and what the motifs, entering as a soft prior, contribute to its MRR. In CQA (\cref{sec:cqa_results}), we ask how \MERITQuery compares with \UltraQuery, so far the only KGFM adapted to CQA.

\subsection{Link prediction}\label{sec:lp_results}
In this subsection, we answer the following research questions:
\textbf{(RQ2)} \textit{How does the zero-shot and fine-tuning MRR of \MERIT compare with that of SoTA KGFMs in the inductive transfer setting?}, and
\textbf{(RQ3)} \textit{How much of the MRR of \MERIT comes from the attention over all relation names, and how much from the motifs?}

\paragraph{Experimental setup.}
We evaluate on the $54$ benchmarks of \citet{DBLP:conf/iclr/0001YM0Z24}, which exclude the three pre-training graphs, FB15k237, WN18RR and CoDEx Medium.
They are grouped into three families by how much of the entity and relation name vocabularies of $\Gi$ already appears in the training graph $\Gtr$ of each benchmark:\footnote{\famone, \famtwo and \famthree are also known as \emph{transductive}, \emph{inductive entity} and \emph{inductive entity and relation}~\citep{DBLP:conf/iclr/0001YM0Z24}, after the original setting each benchmark was created for; \cref{tab:lp-stats} lists the benchmarks in each, with the three pre-training graphs under \famone.} \famone shares both entities and relation names, \famtwo shares relation names but introduces new entities, and \famthree shares neither.
For results that also include the pre-training graphs, see \cref{app:full57}; the conclusions of \cref{fig:1}, over $56$ benchmarks, are unchanged on these $54$ (\cref{tab:basis}).
We compare \MERIT with \ULTRA, \MOTIF, \TRIX and \Flock (\cref{sec:kgfms}) and report the mean and the sample standard deviation over pre-training runs, five for \MERIT, \ULTRA and \TRIX and three for \MOTIF and \Flock (\cref{app:ckpt}).
Following \citet{DBLP:journals/corr/abs-2606-18001}, we evaluate \MOTIF on AristoV4 only once, from its released checkpoint, since its relational hypergraph does not fit on the GPU and must be streamed through a sharded precompute pipeline, and use that value for AristoV4 in all three runs of \MOTIF.
We report MRR and Hits@$10$ under the filtered protocol~\citep{DBLP:conf/nips/BordesUGWY13}, which removes the other known true candidates from the ranking.
We mark in bold the best mean of each column and every mean within one standard deviation of it, its own or that of the best.
We also stratify MRR on the $50$ zero-shot benchmarks of \citet{DBLP:journals/corr/abs-2606-18001} by their four \halflink scenarios, \SQSA, \SQUA, \UQSA and \UQUA, which classify each test triple by whether its query (Q) half $(h,r,?)$ and its answer (A) half $(?,r,t)$ are seen (S) in $\Gi$, that is, witnessed by a triple of $\Ei$, or unseen (U) (\cref{app:ft}).
To keep the inductive transfer setting, we fine-tune per benchmark on \famthree only~\citep{DBLP:journals/corr/abs-2606-18001}, and stratify the fine-tuned MRR by the same \halflink scenarios.
We fine-tune every pre-training run on the training graph $\Gtr$ of each \famthree benchmark, with the fine-tuning protocol of \ULTRA, except for \Flock, which we fine-tune with its own~\citep{DBLP:journals/corr/abs-2510-01510}.
\begin{table}[t]
\centering\scriptsize
\caption{\textbf{Zero-shot \MERIT matches SoTA KGFMs in every family of benchmarks}. Mean MRR and Hits@10 over the benchmarks of each family and over all $54$, averaged over five pre-training runs per model, three for \MOTIF and \Flock, with a single run of \MOTIF on AristoV4; $\pm$ is the sample standard deviation over runs. \textbf{Bold} marks the best mean of each column and every mean within one standard deviation of it, its own or that of the best.}
\label{tab:family-avg}
\resizebox{\linewidth}{!}{%
\begin{tabular}{@{}l cc cc cc cc cc@{}}
\toprule
\textbf{Model} & \multicolumn{2}{c}{\textbf{\famone}} & \multicolumn{2}{c}{\textbf{\famtwo}} & \multicolumn{2}{c}{\textbf{\famthree}} & \multicolumn{2}{c}{\textbf{Total}} \\
 & \multicolumn{2}{c}{(13 graphs)} & \multicolumn{2}{c}{(18 graphs)} & \multicolumn{2}{c}{(23 graphs)} & \multicolumn{2}{c}{(54 graphs)} \\
\cmidrule(lr){2-3} \cmidrule(lr){4-5} \cmidrule(lr){6-7} \cmidrule(lr){8-9}
 & \textbf{MRR} & \textbf{Hits@10} & \textbf{MRR} & \textbf{Hits@10} & \textbf{MRR} & \textbf{Hits@10} & \textbf{MRR} & \textbf{Hits@10} \\
\midrule
\ULTRA & $0.310{\scriptstyle\pm.005}$ & $0.453{\scriptstyle\pm.007}$ & $0.434{\scriptstyle\pm.003}$ & $0.571{\scriptstyle\pm.003}$ & $0.342{\scriptstyle\pm.006}$ & $0.519{\scriptstyle\pm.004}$ & $0.365{\scriptstyle\pm.003}$ & $0.521{\scriptstyle\pm.003}$ \\
\MOTIF & $0.306{\scriptstyle\pm.005}$ & $0.445{\scriptstyle\pm.007}$ & $0.431{\scriptstyle\pm.008}$ & $0.575{\scriptstyle\pm.008}$ & $0.346{\scriptstyle\pm.004}$ & $0.522{\scriptstyle\pm.006}$ & $0.365{\scriptstyle\pm.002}$ & $0.521{\scriptstyle\pm.001}$ \\
\TRIX & $\mathbf{0.329}{\scriptstyle\pm.004}$ & $\mathbf{0.484}{\scriptstyle\pm.005}$ & $\mathbf{0.451}{\scriptstyle\pm.003}$ & $0.589{\scriptstyle\pm.004}$ & $\mathbf{0.362}{\scriptstyle\pm.004}$ & $0.536{\scriptstyle\pm.004}$ & $\mathbf{0.384}{\scriptstyle\pm.002}$ & $0.541{\scriptstyle\pm.003}$ \\
\Flock & $\mathbf{0.331}{\scriptstyle\pm.009}$ & $\mathbf{0.493}{\scriptstyle\pm.011}$ & $\mathbf{0.448}{\scriptstyle\pm.011}$ & $\mathbf{0.602}{\scriptstyle\pm.003}$ & $\mathbf{0.364}{\scriptstyle\pm.007}$ & $\mathbf{0.548}{\scriptstyle\pm.010}$ & $\mathbf{0.384}{\scriptstyle\pm.006}$ & $\mathbf{0.552}{\scriptstyle\pm.005}$ \\
\midrule
\MERIT (ours) & $\mathbf{0.331}{\scriptstyle\pm.005}$ & $\mathbf{0.489}{\scriptstyle\pm.007}$ & $\mathbf{0.454}{\scriptstyle\pm.008}$ & $\mathbf{0.600}{\scriptstyle\pm.005}$ & $\mathbf{0.365}{\scriptstyle\pm.003}$ & $\mathbf{0.544}{\scriptstyle\pm.009}$ & $\mathbf{0.386}{\scriptstyle\pm.002}$ & $\mathbf{0.549}{\scriptstyle\pm.001}$ \\
\bottomrule
\end{tabular}}

\end{table}

\paragraph{\MERIT matches the zero-shot MRR of SoTA KGFMs.}
Over the $54$ benchmarks, \MERIT scores $0.386$ MRR, against $0.384$ for \TRIX and \Flock and $0.365$ for \ULTRA and \MOTIF; in Hits@$10$ it scores $0.549$, within one standard deviation of \Flock at $0.552$ and ahead of \TRIX at $0.541$ (\cref{tab:family-avg}).
Within each family, the mean MRR and Hits@$10$ of \MERIT are the highest or within one standard deviation of the highest (\cref{tab:family-avg}; per benchmark in \cref{app:full57}).
Per \halflink scenario, over the $50$ zero-shot benchmarks, \MERIT has the highest mean on \SQSA, $0.398$ against $0.393$ for \ULTRA, and on \UQSA, $0.568$ against $0.565$ for \Flock, and scores lower than \TRIX and \Flock on \SQUA, $0.175$ against $0.183$ and $0.181$ (\cref{tab:vis50}).
On \UQUA, which isolates generalization beyond $\Gi$~\citep{DBLP:journals/corr/abs-2606-18001}, \MERIT is competitive with \Flock, $0.282$ against $0.276$, and ahead of \ULTRA, \MOTIF and \TRIX at $0.255$ to $0.260$: \MERIT generalizes beyond $\Gi$ as well as \Flock does.

\begin{table}[t]
\noindent
\begin{minipage}[c]{0.63\linewidth}\centering\footnotesize
\setlength{\tabcolsep}{3pt}
\begin{tabular}{@{}l cccc@{}}
\toprule
\textbf{Model} & \textbf{\SQSA} & \textbf{\UQSA} & \textbf{\SQUA} & \textbf{\UQUA} \\
\midrule
\ULTRA & $0.393{\scriptstyle\pm.004}$ & $0.546{\scriptstyle\pm.006}$ & $0.144{\scriptstyle\pm.006}$ & $0.258{\scriptstyle\pm.013}$ \\
\MOTIF & $0.382{\scriptstyle\pm.003}$ & $0.545{\scriptstyle\pm.006}$ & $0.160{\scriptstyle\pm.006}$ & $0.255{\scriptstyle\pm.015}$ \\
\TRIX & $0.392{\scriptstyle\pm.002}$ & $\mathbf{0.564}{\scriptstyle\pm.004}$ & $\mathbf{0.183}{\scriptstyle\pm.004}$ & $0.260{\scriptstyle\pm.014}$ \\
\Flock & $\mathbf{0.389}{\scriptstyle\pm.010}$ & $\mathbf{0.565}{\scriptstyle\pm.008}$ & $\mathbf{0.181}{\scriptstyle\pm.005}$ & $\mathbf{0.276}{\scriptstyle\pm.016}$ \\
\midrule
\MERIT (ours) & $\mathbf{0.398}{\scriptstyle\pm.004}$ & $\mathbf{0.568}{\scriptstyle\pm.008}$ & $0.175{\scriptstyle\pm.005}$ & $\mathbf{0.282}{\scriptstyle\pm.010}$ \\
\bottomrule
\end{tabular}

\end{minipage}\hfill
\begin{minipage}[c]{0.35\linewidth}
\caption{\textbf{The gain of zero-shot \MERIT reaches \UQUA}, and
is on par with \Flock and ahead of the other KGFMs. Mean MRR per \halflink scenario over the $50$ zero-shot benchmarks; averages and bold as in \cref{tab:family-avg}.}
\label{tab:vis50}
\end{minipage}
\end{table}

\begin{figure}[h]
\noindent
\begin{minipage}[c]{0.60\linewidth}\centering
{\figfonts
\colorlet{mMERITft}{olive5}
\colorlet{mULTRAft}{petroil4}
\colorlet{mTRIXft}{bdarker}
\colorlet{mMOTIFft}{gold2}
\colorlet{mFlockft}{lacamlilac}
\pgfplotsset{
  ftaxisV5/.style={
    scale only axis, width=1.32cm, height=2.35cm,
    xmin=-0.35, xmax=1.35, xtick={0,1}, xticklabels={zs,ft},
    xticklabel style={anchor=north, yshift=1pt, text height=1.2ex, text depth=0.2ex},
    axis lines=left, axis line style={-, bgrey1, line width=0.5pt},
    tick style={bgrey1, line width=0.5pt}, grid=major, xmajorgrids=false,
    grid style={bgrey0!22, line width=0.4pt},
    yticklabel style={/pgf/number format/fixed, /pgf/number format/precision=2, /pgf/number format/fixed zerofill},
    scaled y ticks=false, max space between ticks=26pt,
    tick label style={font=\tiny\sffamily, text=bgrey4},
    title style={font=\scriptsize\sffamily, text=bgrey4, at={(rel axis cs:0.05,0.96)}, anchor=north west, inner sep=1pt},
    clip=false,
    every axis plot/.append style={line width=0.8pt},
  },
  sMERITFT/.style={mMERITft, mark=pentagon*, mark size=2.0pt, mark options={fill=mMERITft, draw=white, line width=0.4pt}},
  sULTRAFT/.style={mULTRAft, mark=*, mark size=1.6pt, mark options={fill=mULTRAft, draw=white, line width=0.4pt}},
  sTRIXFT/.style={mTRIXft, mark=triangle*, mark size=1.9pt, mark options={fill=mTRIXft, draw=white, line width=0.4pt}},
  sMOTIFFT/.style={mMOTIFft, mark=square*, mark size=1.5pt, mark options={fill=mMOTIFft, draw=white, line width=0.4pt}},
  sFlockFT/.style={mFlockft, mark=diamond*, mark size=1.9pt, mark options={fill=mFlockft, draw=white, line width=0.4pt}},
}
\includegraphics{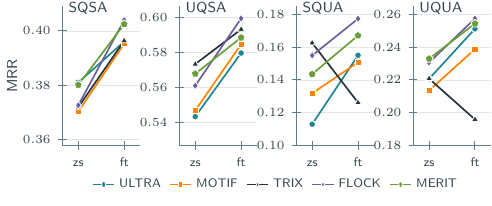}}
\end{minipage}\hfill
\begin{minipage}[c]{0.38\linewidth}
\caption{\textbf{Fine-tuning raises the MRR of \MERIT on every scenario, \UQUA included}. Zero-shot (zs) and fine-tuned (ft) mean MRR over the $22$ \famthree benchmarks, one panel per half-link scenario with its own scale.
}
\label{fig:ft-slope}
\end{minipage}
\end{figure}

\paragraph{Fine-tuning raises \MERIT on every \halflink scenario, while \TRIX loses MRR on the two whose answer half is unseen.}
After per-benchmark fine-tuning, \MERIT gains MRR on every scenario over the $22$ \famthree benchmarks, on \SQUA from $0.144$ to $0.167$ and on \UQUA from $0.233$ to $0.255$, and so do \ULTRA, \MOTIF and \Flock (\cref{fig:ft-slope,tab:ft-family3}).
\TRIX alone loses MRR on these two scenarios, from $0.163$ to $0.126$ on \SQUA and from $0.221$ to $0.196$ on \UQUA.
Over the $22$ benchmarks, fine-tuned \MERIT scores $0.384$ MRR, up from $0.358$ zero-shot, ahead of \ULTRA at $0.375$, \MOTIF at $0.373$ and \TRIX at $0.368$, and slightly below \Flock at $0.391$.

\paragraph{Is attention all you need?}
To answer RQ3, we pre-train two variants of \MERIT, one that keeps only the attention over all relation names and one that keeps only the motifs. The first is the \textit{plain transformer}, with neither the motif bias nor the relation signature, over noise and the query flag (\cref{sec:merit}): it scores $\ablMrrPlain$ MRR over one pre-training run, below \ULTRA at $\ablMrrUltra$. The second is \textit{masked \MERIT}: a relation name $i$ attends only to itself and to the relation names $j$ with ${M_{ij} \neq \mathbf{0}}$, the relation names with which $i$ exchanges messages under motif-constrained message-passing (\cref{sec:kgfms}), and the query-key logit $\smash{a^{(\ell,k)}_{ij}}$ is set to zero, so that its attention depends on the motif bias $\smash{m^k_{ij}}$ alone; it scores $\ablMrrMasked$ over three runs, on par with \ULTRA. With both, \MERIT matches SoTA KGFMs at $\ablMrrMerit$. The motifs help at both points where they enter, the attention and the tokens: removing either the motif bias or the relation signature lowers the MRR of \MERIT (\cref{app:ablations}).

\subsection{Complex query answering}\label{sec:cqa_results}

In this subsection, we answer the following research question:
\textbf{(RQ4)} \textit{How does the MRR of \MERITQuery compare with that of \UltraQuery, the only KGFM for CQA so far, on each family of benchmarks and on different query types?}

\paragraph{Experimental setup.}
The CQA benchmarks of \UltraQuery cover all three families, and their queries come from the sampler of BetaE~\citep{DBLP:conf/nips/RenL20,DBLP:conf/nips/GalkinZR022,DBLP:conf/nips/GalkinZ00Z24}.
\citet{DBLP:conf/icml/GregucciX0LMSV25} show that this sampler skews the benchmarks towards queries simpler than their type suggests: a query of three hops (3p), for instance, can be answered by predicting a single link (1p) when its other two links are already in the graph given to the model, which for KGFMs is the inference graph $\Gi$.
They build the +H benchmarks of \famone, in which the different levels of hardness of a query type are sampled with the same frequency: for example, for a 3p query, they hold the same number of query answer (QA) pairs that \textit{reduce} to predicting a single link, two links, or all three.
We build +H benchmarks for \famtwo and \famthree following their construction, and evaluate on $18$ +H benchmarks: NELL995+H and ICEWS18+H for \famone, and our $9$ \famtwo and $7$ \famthree benchmarks.
We fine-tune \UltraQuery and \MERITQuery on \fbnew with the same recipe, five times each, starting once from each of the five pre-training runs of \ULTRA and of \MERIT (\cref{app:ckpt,app:cqa-details}).
To stay in the inductive transfer setting, we therefore leave FB15k237+H, the third +H benchmark of \famone, out of the $18$ benchmarks of \cref{tab:cqa} and report it only in \cref{app:cqa-types-sec}.
We exclude 1p, which is link prediction, and report the other $15$ query types all together and in three groups: the conjunctive types, whose formula contains only conjunctions, the union types, with a disjunction and no negation, and the negation types, with a negation (\cref{app:cqa-queries}).
We report the MRR of each query type under the filtered protocol of \citet{DBLP:conf/iclr/RenHL20}, and average it over the query types of each group and over the benchmarks.

\begin{table}[t]
\centering\scriptsize
\setlength{\tabcolsep}{3pt}
\caption{\textbf{\MERITQuery is ahead of \UltraQuery over the $18$ benchmarks and on \famone and \famthree, most on the union and negation types}. Mean MRR over the benchmarks of each family and over all $18$, per group of query types, averaged over five fine-tuning runs per method; $\pm$ and bold, within each pair, as in \cref{tab:family-avg}.}
\label{tab:cqa}
\resizebox{\textwidth}{!}{\begin{tabular}{@{}l cc cc cc cc@{}}
\toprule
 & \multicolumn{2}{c}{\textbf{All}} & \multicolumn{2}{c}{\textbf{Conjunctive}} & \multicolumn{2}{c}{\textbf{Union}} & \multicolumn{2}{c}{\textbf{Negation}} \\
\cmidrule(lr){2-3}\cmidrule(lr){4-5}\cmidrule(lr){6-7}\cmidrule(lr){8-9}
\textbf{Family} & \UltraQuery & \MERITQuery & \UltraQuery & \MERITQuery & \UltraQuery & \MERITQuery & \UltraQuery & \MERITQuery \\
\midrule
\famone ($2$) & $0.062{\scriptstyle\pm.012}$ & $\mathbf{0.077{\scriptstyle\pm.005}}$ & $0.080{\scriptstyle\pm.021}$ & $\mathbf{0.103{\scriptstyle\pm.006}}$ & $0.090{\scriptstyle\pm.007}$ & $\mathbf{0.108{\scriptstyle\pm.011}}$ & $\mathbf{0.022{\scriptstyle\pm.003}}$ & $\mathbf{0.023{\scriptstyle\pm.003}}$ \\
\famtwo ($9$) & $\mathbf{0.274{\scriptstyle\pm.003}}$ & $\mathbf{0.276{\scriptstyle\pm.004}}$ & $\mathbf{0.231{\scriptstyle\pm.003}}$ & $0.227{\scriptstyle\pm.003}$ & $\mathbf{0.341{\scriptstyle\pm.012}}$ & $\mathbf{0.350{\scriptstyle\pm.011}}$ & $0.316{\scriptstyle\pm.003}$ & $\mathbf{0.326{\scriptstyle\pm.004}}$ \\
\famthree ($7$) & $0.346{\scriptstyle\pm.003}$ & $\mathbf{0.368{\scriptstyle\pm.013}}$ & $0.357{\scriptstyle\pm.004}$ & $\mathbf{0.367{\scriptstyle\pm.009}}$ & $0.324{\scriptstyle\pm.007}$ & $\mathbf{0.367{\scriptstyle\pm.019}}$ & $0.337{\scriptstyle\pm.006}$ & $\mathbf{0.370{\scriptstyle\pm.018}}$ \\
\midrule
All ($18$) & $0.278{\scriptstyle\pm.003}$ & $\mathbf{0.290{\scriptstyle\pm.006}}$ & $0.263{\scriptstyle\pm.004}$ & $\mathbf{0.268{\scriptstyle\pm.004}}$ & $0.306{\scriptstyle\pm.008}$ & $\mathbf{0.330{\scriptstyle\pm.008}}$ & $0.291{\scriptstyle\pm.003}$ & $\mathbf{0.310{\scriptstyle\pm.009}}$ \\
\bottomrule
\end{tabular}
}
\end{table}

\paragraph{\MERITQuery is ahead of \UltraQuery over the $18$ benchmarks and on \famone and \famthree, most on the union and negation types.}
Over the $18$ benchmarks, \MERITQuery scores $\cqaEpMQAllAll$ MRR, against $\cqaEpUQAllAll$ for \UltraQuery, $\cqaEpMQAllUnion$ against $\cqaEpUQAllUnion$ on the union types, $\cqaEpMQAllNeg$ against $\cqaEpUQAllNeg$ on the negation types and $\cqaEpMQAllConj$ against $\cqaEpUQAllConj$ on the conjunctive ones (\cref{tab:cqa}). Per family, the margin is largest on \famthree, $\cqaEpMQFamThreeAll$ against $\cqaEpUQFamThreeAll$, where \MERITQuery is ahead on each of the $15$ query types (\cref{fig:cqa-types}). On \famone, \MERITQuery scores $\cqaEpMQFamOneAll$ against $\cqaEpUQFamOneAll$, and on \famtwo the two are within one standard deviation, $\cqaEpMQFamTwoAll$ against $\cqaEpUQFamTwoAll$, with \UltraQuery marginally ahead on the conjunctive types, $\cqaEpUQFamTwoConj$ against $\cqaEpMQFamTwoConj$, and \MERITQuery ahead on the negation types, $\cqaEpMQFamTwoNeg$ against $\cqaEpUQFamTwoNeg$. The inference time of \MERITQuery matches that of \UltraQuery, and is well below the time that \TRIX, \MOTIF and \Flock would need if they were adapted to CQA (\cref{sec:cost,fig:qph}).

\begin{takeawaybox}[label={takeaway:results}]
\textbf{Takeaway 2.}
With attention over all relation names and the motifs as a soft prior, \MERIT matches the MRR of SoTA KGFMs in link prediction, and \MERITQuery improves over \UltraQuery in CQA. We therefore advise using the motifs as a prior, not as a constraint.
\end{takeawaybox}

\section{Conclusion}
In this paper, we measure the training and inference costs of SoTA KGFMs and reveal that \TRIX and \Flock gain MRR over \ULTRA at $6\times$ and over $100\times$ its inference time. To reverse this trend, we design \MERIT, the first KGFM with a transformer relation encoder, in which every relation name attends to every other and where the motifs enter as a soft prior. We show that \textit{\textbf{\MERIT matches the zero-shot MRR of SoTA KGFMs at costs comparable with those of the fastest KGFM, \ULTRA}}.
We also show that \MERIT needs both the attention over all relation names and the motifs to reach the zero-shot MRR of SoTA KGFMs (\cref{sec:lp_results,app:ablations}).
After fine-tuning on the $22$ \famthree benchmarks, \MERIT gains MRR on every \halflink scenario, as \ULTRA, \MOTIF and \Flock do, whereas \TRIX loses MRR on the two whose answer half is unseen (\cref{fig:ft-slope}).
Because its cost is comparable with that of \ULTRA, we adapt \MERIT to CQA as \MERITQuery, and build +H benchmarks for \famtwo and \famthree following \citet{DBLP:conf/icml/GregucciX0LMSV25}.
Over the $18$ +H benchmarks, \MERITQuery improves over \UltraQuery at matched cost (\cref{tab:cqa,sec:cost}).

Beyond precomputation, further motif templates leave the overall costs of \MERIT unchanged, so \textit{\textbf{richer motif templates become affordable to study}}, such as the higher-order motifs of \MOTIF, and the motif prior can be combined with the text of relation names, as a prior rather than as the relation graph of SEMMA~\citep{DBLP:conf/emnlp/ArunKNXKVS25}.
How the MRR of \MERIT grows as it is pre-trained on more graphs is left for future work, as is an optimization of the entity encoder.

\subsection*{AI use statement}
In this work, we used generative AI tools to polish the manuscript text, debug, and support us in
coding features that we discussed and subsequently validated. We did not use generative AI tools for
the idea, experiment planning, metric definition, literature review, or other required disclosure tasks.
We have reviewed all AI-assisted work. In particular, we checked LLM-generated code for potential
bugs and misalignment between our ideas and their implementation, and tested it for correctness.
We take responsibility for the final content of this work, including
text, claims, or artifacts produced with the aid of generative AI.

\subsection*{Reproducibility statement}
We describe the architecture and the pre-training of \MERIT in \cref{sec:merit,app:impl}, and the setup of the cost measurements in \cref{sec:cost,app:batch}.
For link prediction, \cref{app:ckpt,app:ft23,app:ablations} give the pre-training runs, the fine-tuning protocol and the ablations of \MERIT.
For complex query answering, \cref{app:plush,app:cqa-details} give the construction of the +H benchmarks and the fine-tuning of \MERITQuery, and \cref{app:stats} gives the statistics of every benchmark.
Our code is available at \url{https://github.com/cgregucci/MERIT}; it merges the codebases of \ULTRA, \MOTIF, \TRIX and \Flock, so that \MERIT and all baselines run from the same scripts.
It also includes the checkpoints of every pre-training run of the five models and of every fine-tuning run of \MERITQuery and \UltraQuery, together with the +H benchmarks of \famtwo and \famthree.

\subsection*{Author contributions}
CG had the initial idea of using motifs as a prior in a transformer rather than as a constraint in message passing, and refined it in discussions with DH. CG designed and ran all the experiments and drew all the figures. AV suggested plotting the Pareto front of MRR against resource consumption for all models. All authors critically revised the paper. AV and SS supervised all phases of the project and gave feedback.

\subsection*{Acknowledgements}
AV was supported by the ``UNREAL: Unified Reasoning Layer for Trustworthy ML'' project (EP/Y023838/1) selected by the ERC and funded by UKRI EPSRC.
CG was funded by the CHIPS Joint Undertaking (JU) under grant agreement No. 101140087 (SMARTY), and by the German Federal Ministry of Education and Research (BMBF) under the sub-project with funding number 16MEE0444.
CG acknowledge compute time on bwHPC and HoreKa HPC (NHR@KIT), funded by the BMBF and the MWK of Baden-Württemberg through the NHR program, with additional support from the DFG. DH was funded by the Deutsche Forschungsgemeinschaft (DFG, German Research Foundation) under Germany's
Excellence Strategy -- EXC 2120/2 -- 390831618.

\bibliography{lp}
\bibliographystyle{iclr2027_conference}

\appendix
\counterwithin{table}{section}
\counterwithin{figure}{section}
\renewcommand{\thetable}{\thesection.\arabic{table}}
\renewcommand{\thefigure}{\thesection.\arabic{figure}}
\clearpage
\section{\MERIT implementation details}\label{app:impl}

\paragraph{Architecture.} The relation transformer has $L = 6$ pre-norm encoder layers with $K = 8$ heads, token dimension $64$ and feed-forward width $256$, without dropout and without a final layer normalization; the motif bias matrix $B \in \mathbb{R}^{4 \times 8}$ and the linear map of the signature, $W_g \in \mathbb{R}^{8 \times 8}$ without bias, are initialized to zero. The entity encoder is the six-layer NBFNet of \ULTRA, with DistMult messages, sum aggregation, residual connections and layer normalization, followed by a two-layer scoring MLP.

\paragraph{Motif tensor.} Let $t \in \{1, \ldots, T\}$, $T = 4$, index the motif templates \hth, \ttt, \htt{} and \tth{} in this order. Our $\Psi$ sets $M_{ijt} = \log(1 + C_{ijt})$, where $C_{ijt}$ is the number of occurrences of motif $t$ between $i$ and $j$, one per distinct entity; we take the logarithm so that counts spanning orders of magnitude do not saturate the attention. For example, in \cref{fig:rel-graph} the \ttt{} motif occurs once between \emph{comprises} and \emph{hasLandmark}, at Colosseo, and the \tth{} motif occurs once, at Rome, so the pair descriptor is $M_{r_1 r_2} = (0, \log 2, 0, \log 2)$ for $r_1 = $ \emph{comprises} and $r_2 = $ \emph{hasLandmark}. We compute $M$ with the four sparse matrix products with which \ULTRA builds its relation graph~\citep{DBLP:conf/iclr/0001YM0Z24}. 

\paragraph{Token layout.} A token has $64$ entries: $8$ hold the signature $\mathrm{s}_r$, $56$ hold $\mathbf{z}_{r \mid p}$, all ones for the query relation and standard Gaussian noise for every other relation name. The noise is redrawn at every forward pass, also at inference, so \MERIT is stochastic at evaluation.

\paragraph{Relation signature.} The fixed summary of the row $M_r$ in $\mathrm{s}_r = W_g\,\phi(M_r)$ is
\[
\phi(M_r) = \big(\, \max_{j \neq r} M_{rj1}, \ldots, \max_{j \neq r} M_{rjT},\ \log(1 + n_1), \ldots, \log(1 + n_T) \,\big),
\]
with $n_t = |\{\, j \neq r : M_{rjt} > 0 \,\}|$, and $W_g\colon \mathbb{R}^{2T} \to \mathbb{R}^{d}$, $d = 8$, is a linear map. A maximum and a count do not depend on the order of the other relation names, so $g$ reads the row as a multiset, as \cref{eq:framework} requires.

\paragraph{Pretraining.} \MERIT is pretrained on the three graphs of \ULTRA, FB15k237, WN18RR and CoDEx Medium, with AdamW, batches of $64$ queries, $512$ strict negatives per query and adversarial temperature $1$. The motif tensor is rebuilt on every training batch from the conditioning graph, and built once per graph at inference.

\paragraph{Inference kernel.} For inference we follow the standard transformer setting~\citep{DBLP:conf/iclr/MicikeviciusNAD18,DBLP:conf/mlsys/PopeDCDBHXAD23} and run the attention of the relation transformer with the FlexAttention kernel~\citep{DBLP:journals/corr/abs-2412-05496}.
The FlexAttention kernel applies the motif bias inside the attention kernel instead of materializing it as an $|\Rin| \times |\Rin|$ matrix per head and per query.
All inference times and FLOPs of \MERIT in the paper are measured on this path, at test batch size $32$.

\clearpage
\section{Cost measurements}\label{app:cost}

\subsection{Batch sizes of the cost measurements}\label{app:batch}
We run all five models from one code base, which merges those of \ULTRA, \MOTIF, \TRIX and \Flock, so that every model runs from the same scripts (\url{https://github.com/cgregucci/MERIT} \cg{placeholder repository, to edit}). We measure the pre-training time per batch of every model at batch size $8$ on one GPU. At inference, we run \ULTRA, \TRIX, \MOTIF and \MERIT at batch size $32$, and \Flock with the zero-shot configurations released with the code of \citet{DBLP:journals/corr/abs-2510-01510}, which set for each benchmark the number of base walks $n$, the number of ensembled passes $P$ per query and the batch size. Since the walks of all queries and all $P$ passes in a batch are held in memory at once, the batch size of \Flock shrinks as $n$ grows: with $n = 512$ and $P = 2$, on CoDEx Large, DBpedia100k and Hetionet, \Flock runs at batch size $2$, and with $n = 16$ and $P = 16$, it runs at batch size $32$ only on the inductive benchmarks and at $16$ on the transductive ones. At the same batch size of $32$, on the $34$ benchmarks of \famtwo and \famthree on which \Flock uses $n = 16$, \MERIT needs $55$ seconds of inference in total against $4.67$ hours for \Flock, which takes $305\times$ the time of \MERIT. We keep these configurations to preserve the performance of \Flock, which \citet{DBLP:journals/corr/abs-2510-01510} show to improve with more ensembled passes and with a number of base walks adapted to the size of the graph.

\subsection{Inference cost by graph size}\label{app:scaling}

\begin{figure}[p]
\centering
{\figfonts
\colorlet{mMERITsa}{olive5}
\colorlet{mULTRAsa}{petroil4}
\colorlet{mTRIXsa}{bdarker}
\colorlet{mMOTIFsa}{gold2}
\colorlet{mFlocksa}{lacamlilac}
\pgfplotsset{
  scalingaxisSA/.style={
    width=5.0cm, height=3.7cm, xmode=log, ymode=log,
    axis lines=left, axis line style={-, bgrey1, line width=0.5pt},
    tick style={bgrey1, line width=0.5pt}, grid=major,
    grid style={bgrey0!22, line width=0.4pt}, log ticks with fixed point,
    tick label style={font=\tiny\sffamily, text=bgrey4},
    label style={font=\tiny\sffamily, text=bgrey4},
    title style={font=\scriptsize\sffamily, text=bgrey4, yshift=-7pt},
    ylabel shift=-4pt,
    enlarge x limits=0.08, clip=false, unbounded coords=discard, xminorticks=false,
    every axis plot/.append style={line width=0.7pt},
  },
  sMERITSA/.style={mMERITsa, mark=pentagon*, mark size=2.2pt, mark options={fill=mMERITsa, draw=white, line width=0.4pt}},
  sULTRASA/.style={mULTRAsa, mark=*, mark size=1.7pt, mark options={fill=mULTRAsa, draw=white, line width=0.4pt}},
  sTRIXSA/.style={mTRIXsa, mark=triangle*, mark size=2.0pt, mark options={fill=mTRIXsa, draw=white, line width=0.4pt}},
  sMOTIFSA/.style={mMOTIFsa, mark=square*, mark size=1.6pt, mark options={fill=mMOTIFsa, draw=white, line width=0.4pt}},
  sFlockSA/.style={mFlocksa, mark=diamond*, mark size=2.0pt, mark options={fill=mFlocksa, draw=white, line width=0.4pt}},
}
\includegraphics{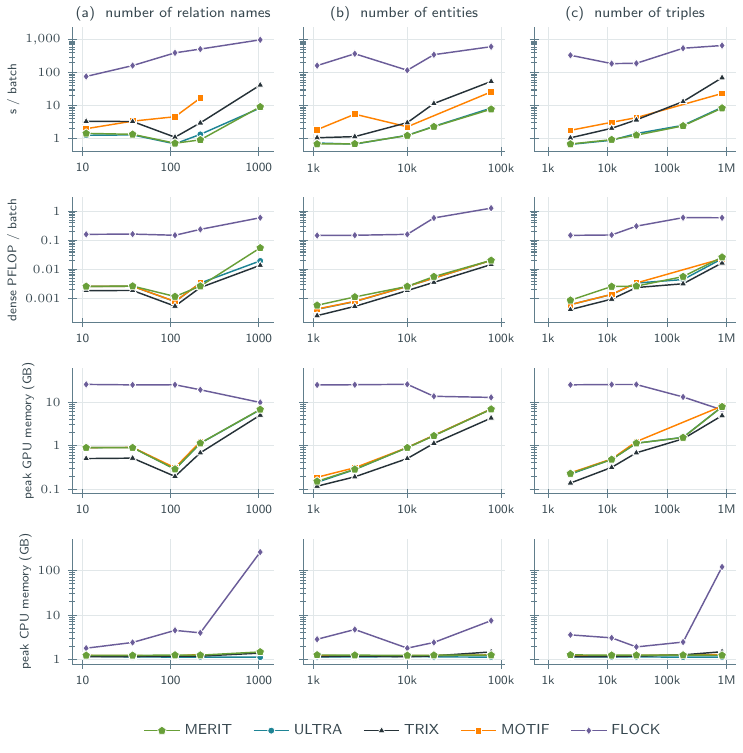}}
\caption{\textbf{\MERIT tracks \ULTRA in wall-clock, GPU memory and CPU memory on every
size axis}, and deviates from it only in dense FLOPs, by the largest factor on the graphs with the
most relation names. Rows show inference wall-clock and FLOPs of dense tensor operators as counted by
PyTorch, both per batch of $1{,}000$ test triples, and peak GPU and CPU memory;
columns bin the benchmarks by relation names, entities and triples of the inference graph
(\cref{tab:scaling-bins}) and markers are medians per bin. \MOTIF markers are omitted from the bins that contain AristoV4, which has no \MOTIF run, so that every marker in a bin is a median over the same benchmarks.
}
\label{fig:scaling-app}
\end{figure}

\cref{fig:scaling} of \cref{sec:cost} bins the benchmarks by the number of relation names and of triples of the inference graph, and plots the median inference time of each model in each bin. \cref{tab:scaling-bins} lists the benchmarks in each bin, and \cref{fig:scaling-app} adds the bins by number of entities and, over all three, the FLOPs, the peak GPU memory and the peak CPU memory at inference.

\begin{table}[!ht]
\centering\scriptsize
\begin{tabular}{@{}llrrp{6.6cm}@{}}
\toprule
axis & bin & $n$ & median & benchmarks\\
\midrule
relation names & $7$ to $20$ & 13 & 11 & Wiki MT1 health, WN v1, WN v4, Wiki MT1 tax, WN v2, HM 1k, HM 3k, HM 5k, WN18RR, WN v3, Wiki MT2 org, NELL v1, Wiki MT2 sci\\
 & $20$ to $60$ & 13 & 37 & Wiki MT4 health, Hetionet, Wiki MT3 infra, Metafam, ConceptNet100k, WK-100, YAGO310, CoDEx Small, Wiki MT4 sci, Wiki MT3 art, ILPC Small, CoDEx Medium, NL-100\\
 & $60$ to $200$ & 17 & 112 & ILPC Large, WK-75, CoDEx Large, WK-25, NELL v4, FB-100, NELL v2, WK-50, NL-0, NL-75, NL-50, NL-25, WDsinger, NELL v3, FB v1, FBNELL, FB-75\\
 & $200$ to $400$ & 12 & 217 & FB v2, NELL23k, NELL995, FB-50, FB v3, FB-25, FB v4, IndigoBM, FB15k237, FB15k237\_10, FB15k237\_20, FB15k237\_50\\
 & $400$ to $2k$ & 2 & 1.04k & DBpedia100k, AristoV4$^\ast$\\
\midrule
entities & $200$ to $2k$ & 7 & 1.09k & NELL v1, Metafam, WN v1, FB v1, NL-75, FB v2, NL-100\\
 & $2k$ to $5k$ & 17 & 2.76k & NL-0, CoDEx Small, NELL v2, NL-25, NL-50, FB v3, FB-100, WK-75, WN v2, FB-75, NELL v4, FB v4, WK-25, NELL v3, FB-25, FB-50, FBNELL\\
 & $5k$ to $12k$ & 15 & 10k & WN v3, ILPC Small, WN v4, WK-50, HM 1k, Wiki MT1 health, Wiki MT1 tax, Wiki MT2 org, Wiki MT2 sci, Wiki MT3 art, Wiki MT3 infra, Wiki MT4 health, Wiki MT4 sci, WDsinger, FB15k237\_10\\
 & $12k$ to $50k$ & 13 & 19.2k & WK-100, FB15k237\_20, FB15k237\_50, FB15k237, IndigoBM, CoDEx Medium, HM 3k, NELL23k, HM 5k, ILPC Large, WN18RR, AristoV4$^\ast$, Hetionet\\
 & $50k$ to $200k$ & 5 & 78.3k & NELL995, CoDEx Large, ConceptNet100k, DBpedia100k, YAGO310\\
\midrule
triples & $800$ to $5k$ & 13 & 2.38k & NELL v1, WN v1, NL-75, FB v1, NL-25, NL-0, NL-100, NL-50, WK-25, WK-75, WN v2, FB v2, NELL v2\\
 & $5k$ to $20k$ & 21 & 11.7k & WN v3, FB-100, NELL v4, Metafam, FB v3, NELL v3, FB-75, WK-50, FBNELL, FB-50, FB v4, WN v4, Wiki MT4 sci, WK-100, Wiki MT1 health, Wiki MT2 sci, Wiki MT4 health, WDsinger, Wiki MT1 tax, FB-25, HM 1k\\
 & $20k$ to $100k$ & 12 & 30.5k & ILPC Small, Wiki MT3 infra, Wiki MT2 org, NELL23k, FB15k237\_10, Wiki MT3 art, CoDEx Small, HM 3k, HM 5k, FB15k237\_20, ILPC Large, WN18RR\\
 & $100k$ to $500k$ & 7 & 186k & ConceptNet100k, FB15k237\_50, NELL995, CoDEx Medium, AristoV4$^\ast$, IndigoBM, FB15k237\\
 & $500k$ to $3M$ & 4 & 838k & CoDEx Large, DBpedia100k, YAGO310, Hetionet\\
\bottomrule
\end{tabular}

\caption{\textbf{Details of the bins behind \cref{fig:scaling,fig:scaling-app}}.
Rows list the benchmarks in each log-spaced bin of relation names, entities and triples of the
inference graph, with the bin count $n$ and the median size. $^\ast$ no \MOTIF run.}
\label{tab:scaling-bins}
\end{table}

\subsection{Inference resources per stage}\label{app:stages}

\begin{figure}[!ht]
\centering
{\figfonts
\colorlet{sPreSB}{bgrey0!65}
\colorlet{sRelSB}{pink3}
\colorlet{sEntSB}{yellow3}
\colorlet{sScoSB}{brown1}
\pgfplotsset{
  stagesaxisSB/.style={
    scale only axis, height=2.05cm,
    symbolic y coords={MOTIF,TRIX,MERIT,ULTRA}, ytick=data,
    xmin=0, enlarge y limits=0.22, enlarge x limits={upper, value=0.04},
    axis lines=left, axis line style={-, bgrey1, line width=0.5pt}, x axis line style={draw=none},
    tick style={bgrey1, line width=0.5pt}, xmajorgrids,
    grid style={bgrey0!22, line width=0.4pt},
    tick label style={font=\tiny\sffamily, text=bgrey4},
    yticklabel style={font=\scriptsize\sffamily, text=bgrey4},
    title style={font=\scriptsize\sffamily, text=bgrey4, yshift=-2pt},
    clip=false,
    every axis plot/.append style={draw=white, line width=0.3pt},
    nodes near coords style={font=\tiny\sffamily, text=bgrey5, anchor=center},
    legend style={font=\scriptsize\sffamily, text=bgrey4, draw=none, /tikz/every even column/.append style={column sep=8pt}},
  },
  stackedSB/.style={xbar stacked, bar width=8pt, point meta=explicit symbolic, nodes near coords, nodes near coords align={center}},
  groupedSB/.style={xbar, bar width=2.6pt, point meta=explicit symbolic},
}
\includegraphics{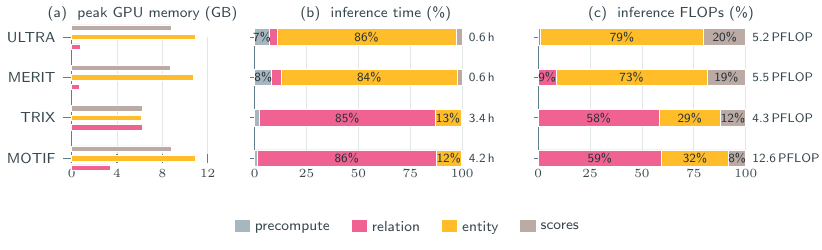}}
\caption{\textbf{The relation stage carries most of the resource consumption of \TRIX and \MOTIF but a
small share for \MERIT and \ULTRA}, whose cost sits in the shared entity GNN. Bars cover the $56$ benchmarks of \cref{fig:1} for the four stages of GNN-based KGFMs:
precomputation of the relational structure, relation encoder, entity encoder and score
computation. Panels show (a) the largest peak GPU memory of each stage on any benchmark,
(b) the share of the wall-clock summed over the suite and (c) the share of the FLOPs summed
over the suite, both with the sum printed at the end of each bar; labels
give the share of the larger segments. Note that \TRIX ties its relation computation to a
small entity GNN, hence its relation-stage peak closely matches that of the entity stage.}
\label{fig:stages}
\end{figure}

For the GNN-based KGFMs, \ULTRA, \MOTIF, \TRIX and \MERIT, we split inference into the four stages of \cref{sec:cost}. \cref{fig:stages} gives the largest peak GPU memory of each stage and its share of the inference time and of the FLOPs over the $56$ benchmarks of \cref{fig:1}.

\subsection{Inference cost with more motif templates}\label{app:tcost}

\begin{figure}[h]
\centering
{\figfonts
\colorlet{cBin0TcV4}{yellow5}
\colorlet{cBin1TcV4}{yellow4}
\colorlet{cBin2TcV4}{gold2}
\colorlet{cBin3TcV4}{tomato2}
\colorlet{cBin4TcV4}{tomato5}
\pgfplotsset{
  tcaxisV4/.style={
    scale only axis, width=4.3cm, height=2.4cm,
    xmin=3.4, xmax=19, xtick={4,8,16}, xticklabels={4,8,16},
    axis lines=left, axis line style={-, bgrey1, line width=0.5pt},
    tick style={bgrey1, line width=0.5pt}, grid=major,
    grid style={bgrey0!22, line width=0.4pt}, log ticks with fixed point, yminorticks=false, xminorticks=false, yticklabel style={/pgf/number format/1000 sep={}},
    tick label style={font=\tiny\sffamily, text=bgrey4},
    label style={font=\tiny\sffamily, text=bgrey4},
    title style={font=\scriptsize\sffamily, text=bgrey4, yshift=-4pt},
    clip=false, unbounded coords=discard,
    every axis plot/.append style={line width=0.8pt},
  },
  sBin0TcV4/.style={cBin0TcV4, mark=*, mark size=1.6pt, mark options={fill=cBin0TcV4, draw=white, line width=0.4pt}},
  sBin1TcV4/.style={cBin1TcV4, mark=*, mark size=1.6pt, mark options={fill=cBin1TcV4, draw=white, line width=0.4pt}},
  sBin2TcV4/.style={cBin2TcV4, mark=*, mark size=1.6pt, mark options={fill=cBin2TcV4, draw=white, line width=0.4pt}},
  sBin3TcV4/.style={cBin3TcV4, mark=*, mark size=1.6pt, mark options={fill=cBin3TcV4, draw=white, line width=0.4pt}},
  sBin4TcV4/.style={cBin4TcV4, mark=*, mark size=1.6pt, mark options={fill=cBin4TcV4, draw=white, line width=0.4pt}},
}
\includegraphics{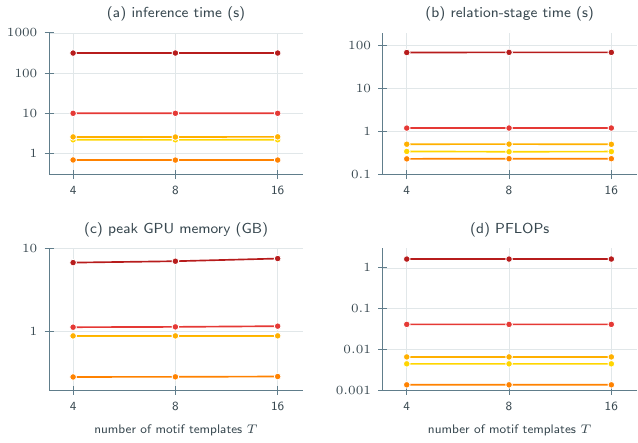}\\[2pt]
\includegraphics{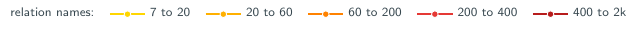}}
\caption{\textbf{The resources of \MERIT stay almost constant as the number of motif templates grows}. As indicated in \cref{sec:cost}, peak GPU memory increases slightly with $T$ on the bins with the most relation names. Median resources of \MERIT over the benchmarks in each relation-name bin of \cref{tab:scaling-bins}, at $T = 4$ and with the four slices of $M$ duplicated to $T = 8$ and $T = 16$. Each line corresponds to a different bin.}
\label{fig:tcost}
\end{figure}

\cref{fig:tcost} breaks down by the number of relation names the measurement of \cref{sec:cost} in which we duplicate the four slices of $M$ to $T = 8$ and $T = 16$: for each relation-name bin of \cref{tab:scaling-bins}, it plots the median resource consumption of \MERIT at $T = 4$, $8$ and $16$.

\subsection{Inference cost per query type}\label{app:qph}
\begin{figure}[!ht]
\noindent
\begin{minipage}[c]{0.60\linewidth}\centering
{\figfonts
\colorlet{mULTRAqh}{petroil4}
\colorlet{mMOTIFqh}{gold2}
\colorlet{mTRIXqh}{bdarker}
\colorlet{mFlockqh}{lacamlilac}
\colorlet{mMERITqh}{olive5}
\pgfplotsset{
  qphaxisQH/.style={
    scale only axis, width=7.9cm, height=3.0cm,
    xmin=\qphXmin, xmax=\qphXmax, ymin=-1.4, ymax=1.4, ytick=\empty,
    xtick={1000,10000,100000,1000000,10000000}, xticklabels={1k,10k,100k,1M,10M},
    xminorticks=true, minor tick style={bgrey1, line width=0.35pt},
    xlabel={queries answered in one H100 hour},
    xlabel style={font=\scriptsize\sffamily, text=bgrey4, yshift=2pt},
    axis lines=left, axis line style={-, bgrey1, line width=0.5pt}, y axis line style={draw=none},
    tick style={bgrey1, line width=0.5pt},
    xmajorgrids, grid style={bgrey0!22, line width=0.4pt},
    tick label style={font=\tiny\sffamily, text=bgrey4},
    nodes near coords style={font=\tiny\sffamily, text=bgrey4, inner sep=1pt},
    clip=false,
  },
  qphAboveQH/.style={only marks, point meta=explicit symbolic, nodes near coords, nodes near coords align={above}, nodes near coords style={yshift=2pt}},
  qphBelowQH/.style={only marks, point meta=explicit symbolic, nodes near coords, nodes near coords align={below}, nodes near coords style={yshift=-2pt}},
  qphULTRAQH/.style={mark=*, mark size=1.6pt, mark options={fill=mULTRAqh, draw=white, line width=0.4pt}},
  qphMOTIFQH/.style={mark=square*, mark size=1.5pt, mark options={fill=mMOTIFqh, draw=white, line width=0.4pt}},
  qphTRIXQH/.style={mark=triangle*, mark size=1.9pt, mark options={fill=mTRIXqh, draw=white, line width=0.4pt}},
  qphFlockQH/.style={mark=diamond*, mark size=1.9pt, mark options={fill=mFlockqh, draw=white, line width=0.4pt}},
  qphMERITQH/.style={mark=pentagon*, mark size=2.0pt, mark options={fill=mMERITqh, draw=white, line width=0.4pt}},
}
\tikzset{qphNameQH/.style={font=\tiny\sffamily, inner sep=1pt, anchor=west, xshift=4pt}}
\includegraphics{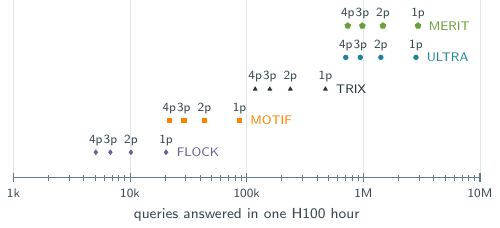}}
\end{minipage}\hfill
\begin{minipage}[c]{0.38\linewidth}
\caption{\textbf{\MERIT answers more \textit{4p} queries in one hour than \TRIX, \MOTIF or \Flock answer \textit{1p} queries, and as many queries as \ULTRA at every hop}. }
\label{fig:qph}
\end{minipage}
\end{figure}

We measure for each model the inference time of link prediction per test query of \fbnew, which is the time of a \textit{1p} query. For \UltraQuery and \MERITQuery, we also measure the time of \textit{2p} and \textit{3p} queries (\cref{sec:cost}), and find that a query with $k$ hops needs about $k$ times the time of a \textit{1p} query. We assume that the time of \TRIX, \MOTIF and \Flock, if adapted as \ULTRA is in \UltraQuery, would grow with the number of hops in the same way. \cref{fig:qph} shows the resulting projection of the number of path queries each KGFM answers in one hour: \ULTRA and \MERIT would answer more \textit{4p} queries in one hour than the other KGFMs answer \textit{1p} queries.

\clearpage
\section{Link prediction}\label{app:lp}

\subsection{Pre-trained checkpoints}\label{app:ckpt}

The mean of every SoTA KGFM is over the checkpoint released by its authors and further pre-training runs from fresh random seeds.
For \ULTRA, \MOTIF and \TRIX we use the pre-training runs of \citet{DBLP:journals/corr/abs-2606-18001}: the released checkpoint and four seeds for \ULTRA and \TRIX, the released checkpoint and two seeds for \MOTIF.
For \Flock, the official repository releases a single checkpoint, so we pre-train two further seeds ourselves, for three runs of \Flock.
\MERIT has five pre-training runs, one per seed.

\clearpage
\subsection{Zero-shot link prediction per benchmark}\label{app:full57}

\cref{tab:lp-bases} adds to the $54$ benchmarks of \cref{sec:lp_results} the mean over all $57$ benchmarks, and \cref{tab:full57-mrr,tab:full57-hits} give the MRR and Hits@10 on each of the $57$.
\Flock samples random walks at inference and is not seeded there, so its sample standard deviation carries both pre-training and inference variability.
\MERIT redraws its noise at every forward pass, also at inference (\cref{app:impl}), but this variability is negligible against that over pre-training runs: over five evaluation seeds of one pre-training run, its mean MRR over the $57$ benchmarks has a sample standard deviation of $0.0003$, against $0.0024$ over the five pre-training runs.
\cref{tab:basis} compares inference time and MRR on the $54$ benchmarks, and shows that the conclusions of \cref{fig:1}, over the $56$ benchmarks other than AristoV4 (\cref{sec:cost}), are unchanged on these $54$.

\begin{table}[!ht]
\centering\scriptsize
\caption{\textbf{Zero-shot MRR and Hits@10 of \MERIT and the SoTA KGFMs on the $54$ benchmarks and on all $57$}. Mean over the benchmarks of each column, averaged over five pre-training runs per model, three for \MOTIF and \Flock; $\pm$ is the sample standard deviation over runs. The single run of \MOTIF on AristoV4 enters every run of \MOTIF. Bold as in \cref{tab:family-avg}.}
\label{tab:lp-bases}
\small
\begin{tabular}{@{}l cc cc@{}}
\toprule
\textbf{Model} & \multicolumn{2}{c}{\textbf{54 benchmarks}} & \multicolumn{2}{c}{\textbf{57 benchmarks}} \\
\cmidrule(lr){2-3} \cmidrule(lr){4-5}
 & \textbf{MRR} & \textbf{Hits@10} & \textbf{MRR} & \textbf{Hits@10} \\
\midrule
\ULTRA & $0.365{\scriptstyle\pm.003}$ & $0.521{\scriptstyle\pm.003}$ & $0.368{\scriptstyle\pm.002}$ & $0.523{\scriptstyle\pm.004}$ \\
\MOTIF & $0.365{\scriptstyle\pm.002}$ & $0.521{\scriptstyle\pm.001}$ & $0.368{\scriptstyle\pm.001}$ & $0.524{\scriptstyle\pm.001}$ \\
\TRIX & $\mathbf{0.384}{\scriptstyle\pm.002}$ & $0.541{\scriptstyle\pm.003}$ & $0.386{\scriptstyle\pm.002}$ & $0.542{\scriptstyle\pm.003}$ \\
\Flock & $\mathbf{0.384}{\scriptstyle\pm.006}$ & $\mathbf{0.552}{\scriptstyle\pm.005}$ & $0.382{\scriptstyle\pm.005}$ & $\mathbf{0.548}{\scriptstyle\pm.005}$ \\
\midrule
\MERIT (ours) & $\mathbf{0.386}{\scriptstyle\pm.002}$ & $\mathbf{0.549}{\scriptstyle\pm.001}$ & $\mathbf{0.388}{\scriptstyle\pm.002}$ & $\mathbf{0.551}{\scriptstyle\pm.001}$ \\
\bottomrule
\end{tabular}

\end{table}

\begin{table}[!ht]
\noindent
\begin{minipage}[c]{0.56\linewidth}\centering\small
\begin{tabular}{@{}l cc@{}}
\toprule
\textbf{Model} & \textbf{inference time (h)} & \textbf{MRR} \\
\midrule
\ULTRA & $0.60$ & $0.365{\scriptstyle\pm.003}$ \\
\MOTIF & $3.69$$^\ast$ & $0.365{\scriptstyle\pm.002}$ \\
\TRIX & $3.31$ & $0.384{\scriptstyle\pm.002}$ \\
\Flock & $67.07$ & $0.384{\scriptstyle\pm.006}$ \\
\midrule
\MERIT (ours) & $0.58$ & $0.386{\scriptstyle\pm.002}$ \\
\bottomrule
\end{tabular}

\end{minipage}\hfill
\begin{minipage}[c]{0.41\linewidth}
\caption{\textbf{The conclusions of \cref{fig:1} are unchanged on the $54$ benchmarks in the inductive transfer setting}. Inference time over the $54$ benchmarks in hours and MRR averaged over five pre-training runs per model, three for \MOTIF and \Flock; $\pm$ is the sample standard deviation over runs. The single run of \MOTIF on AristoV4 enters every run of \MOTIF, and $^\ast$ marks its time over the $53$ benchmarks other than AristoV4.}
\label{tab:basis}
\end{minipage}
\end{table}

\begin{table}[!ht]
\centering\scriptsize
\caption{\textbf{Zero-shot MRR of \MERIT and the SoTA KGFMs on each of the $57$ benchmarks}. Mean MRR over five pre-training runs per model, three for \MOTIF and \Flock; $\pm$ is the sample standard deviation over runs; $^\ast$ marks the single run of \MOTIF on AristoV4, which enters every run of \MOTIF in the last row, the average over all $57$ benchmarks. \textbf{Bold} marks the best mean of each row.}
\label{tab:full57-mrr}
\begin{tabular}{@{}l ccccc@{}}
\toprule
\textbf{Benchmark} & \textbf{\ULTRA} & \textbf{\MOTIF} & \textbf{\TRIX} & \textbf{\Flock} & \textbf{\MERIT} \\
\midrule
\multicolumn{6}{@{}l}{\textit{\famone (16)}} \\
\midrule
CoDEx Small & $0.467{\scriptstyle\pm.011}$ & $0.476{\scriptstyle\pm.005}$ & $0.478{\scriptstyle\pm.006}$ & $0.446{\scriptstyle\pm.006}$ & $\mathbf{0.479}{\scriptstyle\pm.007}$ \\
CoDEx Medium & $0.368{\scriptstyle\pm.008}$ & $0.369{\scriptstyle\pm.005}$ & $0.364{\scriptstyle\pm.003}$ & $0.267{\scriptstyle\pm.007}$ & $\mathbf{0.374}{\scriptstyle\pm.002}$ \\
CoDEx Large & $0.331{\scriptstyle\pm.004}$ & $0.342{\scriptstyle\pm.004}$ & $0.338{\scriptstyle\pm.002}$ & $0.341{\scriptstyle\pm.002}$ & $\mathbf{0.346}{\scriptstyle\pm.002}$ \\
WDsinger & $0.385{\scriptstyle\pm.017}$ & $0.395{\scriptstyle\pm.003}$ & $0.389{\scriptstyle\pm.012}$ & $\mathbf{0.403}{\scriptstyle\pm.010}$ & $\mathbf{0.403}{\scriptstyle\pm.012}$ \\
FB15k237\_10 & $0.237{\scriptstyle\pm.011}$ & $0.241{\scriptstyle\pm.005}$ & $0.239{\scriptstyle\pm.008}$ & $0.216{\scriptstyle\pm.047}$ & $\mathbf{0.248}{\scriptstyle\pm.002}$ \\
FB15k237\_20 & $0.267{\scriptstyle\pm.004}$ & $0.264{\scriptstyle\pm.005}$ & $0.265{\scriptstyle\pm.004}$ & $0.244{\scriptstyle\pm.049}$ & $\mathbf{0.270}{\scriptstyle\pm.001}$ \\
FB15k237\_50 & $0.321{\scriptstyle\pm.002}$ & $0.317{\scriptstyle\pm.004}$ & $0.322{\scriptstyle\pm.001}$ & $0.290{\scriptstyle\pm.054}$ & $\mathbf{0.326}{\scriptstyle\pm.001}$ \\
FB15k237 & $0.368{\scriptstyle\pm.008}$ & $0.358{\scriptstyle\pm.010}$ & $0.365{\scriptstyle\pm.002}$ & $0.267{\scriptstyle\pm.008}$ & $\mathbf{0.377}{\scriptstyle\pm.002}$ \\
NELL23k & $0.238{\scriptstyle\pm.006}$ & $0.224{\scriptstyle\pm.005}$ & $0.228{\scriptstyle\pm.008}$ & $0.219{\scriptstyle\pm.014}$ & $\mathbf{0.249}{\scriptstyle\pm.006}$ \\
WN18RR & $0.497{\scriptstyle\pm.022}$ & $0.532{\scriptstyle\pm.012}$ & $0.512{\scriptstyle\pm.004}$ & $0.498{\scriptstyle\pm.012}$ & $\mathbf{0.535}{\scriptstyle\pm.002}$ \\
AristoV4 & $0.181{\scriptstyle\pm.016}$ & $0.096^{\ast}$ & $0.190{\scriptstyle\pm.009}$ & $\mathbf{0.300}{\scriptstyle\pm.007}$ & $0.186{\scriptstyle\pm.053}$ \\
Hetionet & $0.237{\scriptstyle\pm.034}$ & $\mathbf{0.273}{\scriptstyle\pm.018}$ & $0.272{\scriptstyle\pm.014}$ & $0.245{\scriptstyle\pm.003}$ & $0.237{\scriptstyle\pm.021}$ \\
NELL995 & $0.453{\scriptstyle\pm.021}$ & $0.478{\scriptstyle\pm.017}$ & $0.474{\scriptstyle\pm.012}$ & $0.482{\scriptstyle\pm.010}$ & $\mathbf{0.510}{\scriptstyle\pm.019}$ \\
ConceptNet100k & $0.069{\scriptstyle\pm.021}$ & $0.037{\scriptstyle\pm.024}$ & $0.217{\scriptstyle\pm.024}$ & $0.238{\scriptstyle\pm.029}$ & $\mathbf{0.251}{\scriptstyle\pm.019}$ \\
DBpedia100k & $0.396{\scriptstyle\pm.013}$ & $0.398{\scriptstyle\pm.014}$ & $0.433{\scriptstyle\pm.008}$ & $\mathbf{0.451}{\scriptstyle\pm.007}$ & $0.415{\scriptstyle\pm.012}$ \\
YAGO310 & $\mathbf{0.447}{\scriptstyle\pm.021}$ & $0.433{\scriptstyle\pm.021}$ & $0.439{\scriptstyle\pm.036}$ & $0.422{\scriptstyle\pm.036}$ & $0.384{\scriptstyle\pm.050}$ \\
\midrule
\multicolumn{6}{@{}l}{\textit{\famtwo (18)}} \\
\midrule
FB v1 & $0.505{\scriptstyle\pm.011}$ & $0.492{\scriptstyle\pm.015}$ & $0.510{\scriptstyle\pm.011}$ & $0.492{\scriptstyle\pm.011}$ & $\mathbf{0.531}{\scriptstyle\pm.007}$ \\
FB v2 & $0.513{\scriptstyle\pm.009}$ & $0.507{\scriptstyle\pm.005}$ & $0.519{\scriptstyle\pm.010}$ & $0.538{\scriptstyle\pm.006}$ & $\mathbf{0.542}{\scriptstyle\pm.004}$ \\
FB v3 & $0.489{\scriptstyle\pm.010}$ & $0.490{\scriptstyle\pm.008}$ & $0.494{\scriptstyle\pm.005}$ & $0.511{\scriptstyle\pm.007}$ & $\mathbf{0.512}{\scriptstyle\pm.004}$ \\
FB v4 & $0.482{\scriptstyle\pm.006}$ & $0.485{\scriptstyle\pm.003}$ & $0.486{\scriptstyle\pm.004}$ & $0.501{\scriptstyle\pm.002}$ & $\mathbf{0.502}{\scriptstyle\pm.002}$ \\
WN v1 & $0.675{\scriptstyle\pm.014}$ & $0.673{\scriptstyle\pm.007}$ & $\mathbf{0.700}{\scriptstyle\pm.007}$ & $0.655{\scriptstyle\pm.053}$ & $0.689{\scriptstyle\pm.033}$ \\
WN v2 & $0.674{\scriptstyle\pm.002}$ & $0.659{\scriptstyle\pm.004}$ & $0.685{\scriptstyle\pm.004}$ & $0.647{\scriptstyle\pm.059}$ & $\mathbf{0.696}{\scriptstyle\pm.008}$ \\
WN v3 & $0.385{\scriptstyle\pm.005}$ & $0.425{\scriptstyle\pm.005}$ & $0.433{\scriptstyle\pm.010}$ & $0.458{\scriptstyle\pm.025}$ & $\mathbf{0.460}{\scriptstyle\pm.011}$ \\
WN v4 & $0.635{\scriptstyle\pm.003}$ & $0.632{\scriptstyle\pm.008}$ & $0.649{\scriptstyle\pm.001}$ & $0.649{\scriptstyle\pm.004}$ & $\mathbf{0.655}{\scriptstyle\pm.004}$ \\
NELL v1 & $0.695{\scriptstyle\pm.053}$ & $0.677{\scriptstyle\pm.106}$ & $\mathbf{0.798}{\scriptstyle\pm.009}$ & $0.675{\scriptstyle\pm.017}$ & $0.632{\scriptstyle\pm.113}$ \\
NELL v2 & $0.528{\scriptstyle\pm.016}$ & $0.547{\scriptstyle\pm.015}$ & $0.553{\scriptstyle\pm.009}$ & $0.565{\scriptstyle\pm.024}$ & $\mathbf{0.570}{\scriptstyle\pm.015}$ \\
NELL v3 & $0.527{\scriptstyle\pm.022}$ & $0.502{\scriptstyle\pm.031}$ & $0.546{\scriptstyle\pm.011}$ & $\mathbf{0.578}{\scriptstyle\pm.019}$ & $0.571{\scriptstyle\pm.012}$ \\
NELL v4 & $0.492{\scriptstyle\pm.014}$ & $0.479{\scriptstyle\pm.022}$ & $0.536{\scriptstyle\pm.007}$ & $\mathbf{0.543}{\scriptstyle\pm.011}$ & $0.533{\scriptstyle\pm.007}$ \\
ILPC Small & $0.294{\scriptstyle\pm.006}$ & $0.296{\scriptstyle\pm.001}$ & $0.301{\scriptstyle\pm.002}$ & $\mathbf{0.309}{\scriptstyle\pm.001}$ & $0.303{\scriptstyle\pm.003}$ \\
ILPC Large & $0.296{\scriptstyle\pm.003}$ & $0.288{\scriptstyle\pm.004}$ & $0.301{\scriptstyle\pm.009}$ & $\mathbf{0.318}{\scriptstyle\pm.001}$ & $0.306{\scriptstyle\pm.004}$ \\
HM 1k & $0.075{\scriptstyle\pm.007}$ & $0.070{\scriptstyle\pm.007}$ & $0.066{\scriptstyle\pm.007}$ & $0.067{\scriptstyle\pm.001}$ & $\mathbf{0.081}{\scriptstyle\pm.005}$ \\
HM 3k & $0.062{\scriptstyle\pm.003}$ & $0.057{\scriptstyle\pm.003}$ & $0.063{\scriptstyle\pm.005}$ & $0.067{\scriptstyle\pm.003}$ & $\mathbf{0.071}{\scriptstyle\pm.001}$ \\
HM 5k & $0.056{\scriptstyle\pm.003}$ & $0.052{\scriptstyle\pm.004}$ & $0.059{\scriptstyle\pm.004}$ & $\mathbf{0.067}{\scriptstyle\pm.003}$ & $\mathbf{0.067}{\scriptstyle\pm.000}$ \\
IndigoBM & $0.438{\scriptstyle\pm.007}$ & $0.434{\scriptstyle\pm.007}$ & $0.426{\scriptstyle\pm.008}$ & $0.422{\scriptstyle\pm.005}$ & $\mathbf{0.447}{\scriptstyle\pm.002}$ \\
\midrule
\multicolumn{6}{@{}l}{\textit{\famthree (23)}} \\
\midrule
FB-25 & $0.391{\scriptstyle\pm.004}$ & $0.386{\scriptstyle\pm.002}$ & $0.383{\scriptstyle\pm.006}$ & $\mathbf{0.402}{\scriptstyle\pm.001}$ & $0.397{\scriptstyle\pm.004}$ \\
FB-50 & $0.340{\scriptstyle\pm.005}$ & $0.339{\scriptstyle\pm.001}$ & $0.315{\scriptstyle\pm.013}$ & $\mathbf{0.352}{\scriptstyle\pm.003}$ & $0.338{\scriptstyle\pm.007}$ \\
FB-75 & $0.401{\scriptstyle\pm.004}$ & $0.394{\scriptstyle\pm.005}$ & $0.392{\scriptstyle\pm.006}$ & $\mathbf{0.416}{\scriptstyle\pm.005}$ & $0.399{\scriptstyle\pm.005}$ \\
FB-100 & $0.436{\scriptstyle\pm.006}$ & $0.423{\scriptstyle\pm.006}$ & $0.437{\scriptstyle\pm.005}$ & $\mathbf{0.452}{\scriptstyle\pm.002}$ & $0.439{\scriptstyle\pm.003}$ \\
WK-25 & $0.297{\scriptstyle\pm.008}$ & $\mathbf{0.312}{\scriptstyle\pm.005}$ & $0.304{\scriptstyle\pm.005}$ & $0.273{\scriptstyle\pm.010}$ & $0.305{\scriptstyle\pm.007}$ \\
WK-50 & $0.159{\scriptstyle\pm.010}$ & $\mathbf{0.165}{\scriptstyle\pm.002}$ & $\mathbf{0.165}{\scriptstyle\pm.004}$ & $0.128{\scriptstyle\pm.013}$ & $0.162{\scriptstyle\pm.011}$ \\
WK-75 & $\mathbf{0.379}{\scriptstyle\pm.006}$ & $0.367{\scriptstyle\pm.001}$ & $0.369{\scriptstyle\pm.005}$ & $0.377{\scriptstyle\pm.008}$ & $\mathbf{0.379}{\scriptstyle\pm.011}$ \\
WK-100 & $0.172{\scriptstyle\pm.005}$ & $0.169{\scriptstyle\pm.010}$ & $0.182{\scriptstyle\pm.005}$ & $\mathbf{0.184}{\scriptstyle\pm.004}$ & $0.182{\scriptstyle\pm.010}$ \\
NL-0 & $0.355{\scriptstyle\pm.009}$ & $0.324{\scriptstyle\pm.004}$ & $0.373{\scriptstyle\pm.011}$ & $0.376{\scriptstyle\pm.013}$ & $\mathbf{0.377}{\scriptstyle\pm.011}$ \\
NL-25 & $0.356{\scriptstyle\pm.026}$ & $0.361{\scriptstyle\pm.021}$ & $\mathbf{0.389}{\scriptstyle\pm.011}$ & $0.339{\scriptstyle\pm.014}$ & $0.375{\scriptstyle\pm.022}$ \\
NL-50 & $0.379{\scriptstyle\pm.014}$ & $0.372{\scriptstyle\pm.011}$ & $0.386{\scriptstyle\pm.016}$ & $0.341{\scriptstyle\pm.024}$ & $\mathbf{0.389}{\scriptstyle\pm.029}$ \\
NL-75 & $0.330{\scriptstyle\pm.021}$ & $0.321{\scriptstyle\pm.010}$ & $\mathbf{0.346}{\scriptstyle\pm.004}$ & $0.291{\scriptstyle\pm.025}$ & $0.313{\scriptstyle\pm.034}$ \\
NL-100 & $0.453{\scriptstyle\pm.016}$ & $0.444{\scriptstyle\pm.005}$ & $\mathbf{0.461}{\scriptstyle\pm.010}$ & $0.429{\scriptstyle\pm.030}$ & $0.456{\scriptstyle\pm.024}$ \\
Metafam & $0.349{\scriptstyle\pm.089}$ & $0.374{\scriptstyle\pm.029}$ & $0.320{\scriptstyle\pm.028}$ & $0.475{\scriptstyle\pm.023}$ & $\mathbf{0.504}{\scriptstyle\pm.075}$ \\
FBNELL & $0.480{\scriptstyle\pm.014}$ & $0.457{\scriptstyle\pm.011}$ & $0.474{\scriptstyle\pm.017}$ & $0.492{\scriptstyle\pm.019}$ & $\mathbf{0.495}{\scriptstyle\pm.014}$ \\
Wiki MT1 tax & $0.234{\scriptstyle\pm.006}$ & $0.290{\scriptstyle\pm.038}$ & $\mathbf{0.343}{\scriptstyle\pm.029}$ & $0.283{\scriptstyle\pm.012}$ & $0.342{\scriptstyle\pm.049}$ \\
Wiki MT1 health & $0.271{\scriptstyle\pm.039}$ & $0.299{\scriptstyle\pm.062}$ & $0.373{\scriptstyle\pm.005}$ & $\mathbf{0.384}{\scriptstyle\pm.001}$ & $0.376{\scriptstyle\pm.006}$ \\
Wiki MT2 org & $0.085{\scriptstyle\pm.002}$ & $0.084{\scriptstyle\pm.008}$ & $0.093{\scriptstyle\pm.003}$ & $\mathbf{0.100}{\scriptstyle\pm.001}$ & $0.094{\scriptstyle\pm.004}$ \\
Wiki MT2 sci & $0.261{\scriptstyle\pm.011}$ & $0.270{\scriptstyle\pm.039}$ & $0.317{\scriptstyle\pm.007}$ & $0.314{\scriptstyle\pm.003}$ & $\mathbf{0.319}{\scriptstyle\pm.024}$ \\
Wiki MT3 art & $0.254{\scriptstyle\pm.004}$ & $0.270{\scriptstyle\pm.004}$ & $0.292{\scriptstyle\pm.009}$ & $\mathbf{0.302}{\scriptstyle\pm.003}$ & $0.293{\scriptstyle\pm.003}$ \\
Wiki MT3 infra & $0.623{\scriptstyle\pm.016}$ & $0.632{\scriptstyle\pm.024}$ & $0.646{\scriptstyle\pm.007}$ & $\mathbf{0.676}{\scriptstyle\pm.006}$ & $0.613{\scriptstyle\pm.045}$ \\
Wiki MT4 sci & $0.284{\scriptstyle\pm.009}$ & $0.273{\scriptstyle\pm.010}$ & $0.287{\scriptstyle\pm.003}$ & $\mathbf{0.298}{\scriptstyle\pm.010}$ & $0.289{\scriptstyle\pm.011}$ \\
Wiki MT4 health & $0.589{\scriptstyle\pm.053}$ & $0.639{\scriptstyle\pm.011}$ & $\mathbf{0.685}{\scriptstyle\pm.006}$ & $0.684{\scriptstyle\pm.009}$ & $0.549{\scriptstyle\pm.064}$ \\
\midrule
Average ($57$) & $0.368{\scriptstyle\pm.002}$ & $0.368{\scriptstyle\pm.001}$ & $0.386{\scriptstyle\pm.002}$ & $0.382{\scriptstyle\pm.005}$ & $\mathbf{0.388}{\scriptstyle\pm.002}$ \\
\bottomrule
\end{tabular}

\end{table}

\begin{table}[!ht]
\centering\scriptsize
\caption{\textbf{Zero-shot Hits@10 of \MERIT and the SoTA KGFMs on each of the $57$ benchmarks}. Mean Hits@10 over five pre-training runs per model, three for \MOTIF and \Flock; $\pm$ is the sample standard deviation over runs; $^\ast$ marks the single run of \MOTIF on AristoV4, which enters every run of \MOTIF in the last row, the average over all $57$ benchmarks. \textbf{Bold} marks the best mean of each row.}
\label{tab:full57-hits}
\begin{tabular}{@{}l ccccc@{}}
\toprule
\textbf{Benchmark} & \textbf{\ULTRA} & \textbf{\MOTIF} & \textbf{\TRIX} & \textbf{\Flock} & \textbf{\MERIT} \\
\midrule
\multicolumn{6}{@{}l}{\textit{\famone (16)}} \\
\midrule
CoDEx Small & $0.664{\scriptstyle\pm.007}$ & $\mathbf{0.673}{\scriptstyle\pm.004}$ & $\mathbf{0.673}{\scriptstyle\pm.003}$ & $0.642{\scriptstyle\pm.004}$ & $0.671{\scriptstyle\pm.004}$ \\
CoDEx Medium & $0.520{\scriptstyle\pm.016}$ & $0.519{\scriptstyle\pm.005}$ & $0.521{\scriptstyle\pm.002}$ & $0.399{\scriptstyle\pm.009}$ & $\mathbf{0.531}{\scriptstyle\pm.003}$ \\
CoDEx Large & $0.461{\scriptstyle\pm.008}$ & $0.471{\scriptstyle\pm.006}$ & $0.472{\scriptstyle\pm.002}$ & $0.465{\scriptstyle\pm.003}$ & $\mathbf{0.479}{\scriptstyle\pm.003}$ \\
WDsinger & $0.496{\scriptstyle\pm.015}$ & $0.512{\scriptstyle\pm.003}$ & $0.499{\scriptstyle\pm.014}$ & $\mathbf{0.526}{\scriptstyle\pm.008}$ & $0.513{\scriptstyle\pm.007}$ \\
FB15k237\_10 & $0.383{\scriptstyle\pm.014}$ & $0.390{\scriptstyle\pm.005}$ & $0.385{\scriptstyle\pm.012}$ & $0.360{\scriptstyle\pm.069}$ & $\mathbf{0.400}{\scriptstyle\pm.003}$ \\
FB15k237\_20 & $0.427{\scriptstyle\pm.005}$ & $0.427{\scriptstyle\pm.004}$ & $0.425{\scriptstyle\pm.006}$ & $0.404{\scriptstyle\pm.069}$ & $\mathbf{0.433}{\scriptstyle\pm.002}$ \\
FB15k237\_50 & $0.520{\scriptstyle\pm.002}$ & $0.514{\scriptstyle\pm.007}$ & $0.520{\scriptstyle\pm.001}$ & $0.480{\scriptstyle\pm.072}$ & $\mathbf{0.526}{\scriptstyle\pm.002}$ \\
FB15k237 & $0.559{\scriptstyle\pm.012}$ & $0.555{\scriptstyle\pm.010}$ & $0.559{\scriptstyle\pm.003}$ & $0.421{\scriptstyle\pm.013}$ & $\mathbf{0.572}{\scriptstyle\pm.002}$ \\
NELL23k & $0.402{\scriptstyle\pm.009}$ & $0.386{\scriptstyle\pm.002}$ & $0.381{\scriptstyle\pm.012}$ & $0.377{\scriptstyle\pm.018}$ & $\mathbf{0.417}{\scriptstyle\pm.005}$ \\
WN18RR & $0.593{\scriptstyle\pm.031}$ & $0.634{\scriptstyle\pm.011}$ & $0.608{\scriptstyle\pm.002}$ & $0.602{\scriptstyle\pm.009}$ & $\mathbf{0.642}{\scriptstyle\pm.001}$ \\
AristoV4 & $0.260{\scriptstyle\pm.024}$ & $0.142^{\ast}$ & $0.289{\scriptstyle\pm.012}$ & $\mathbf{0.431}{\scriptstyle\pm.010}$ & $0.284{\scriptstyle\pm.093}$ \\
Hetionet & $0.342{\scriptstyle\pm.046}$ & $0.397{\scriptstyle\pm.012}$ & $\mathbf{0.404}{\scriptstyle\pm.015}$ & $0.372{\scriptstyle\pm.001}$ & $0.361{\scriptstyle\pm.028}$ \\
NELL995 & $0.595{\scriptstyle\pm.021}$ & $0.621{\scriptstyle\pm.001}$ & $0.625{\scriptstyle\pm.010}$ & $0.634{\scriptstyle\pm.020}$ & $\mathbf{0.641}{\scriptstyle\pm.008}$ \\
ConceptNet100k & $0.137{\scriptstyle\pm.042}$ & $0.078{\scriptstyle\pm.063}$ & $0.380{\scriptstyle\pm.047}$ & $0.438{\scriptstyle\pm.052}$ & $\mathbf{0.441}{\scriptstyle\pm.028}$ \\
DBpedia100k & $0.576{\scriptstyle\pm.006}$ & $0.569{\scriptstyle\pm.007}$ & $0.602{\scriptstyle\pm.004}$ & $\mathbf{0.625}{\scriptstyle\pm.002}$ & $0.605{\scriptstyle\pm.003}$ \\
YAGO310 & $0.620{\scriptstyle\pm.024}$ & $0.604{\scriptstyle\pm.026}$ & $0.640{\scriptstyle\pm.021}$ & $\mathbf{0.655}{\scriptstyle\pm.018}$ & $0.588{\scriptstyle\pm.065}$ \\
\midrule
\multicolumn{6}{@{}l}{\textit{\famtwo (18)}} \\
\midrule
FB v1 & $0.680{\scriptstyle\pm.015}$ & $0.678{\scriptstyle\pm.013}$ & $0.676{\scriptstyle\pm.010}$ & $0.695{\scriptstyle\pm.006}$ & $\mathbf{0.698}{\scriptstyle\pm.009}$ \\
FB v2 & $0.715{\scriptstyle\pm.012}$ & $0.713{\scriptstyle\pm.003}$ & $0.723{\scriptstyle\pm.007}$ & $\mathbf{0.738}{\scriptstyle\pm.002}$ & $0.733{\scriptstyle\pm.004}$ \\
FB v3 & $0.658{\scriptstyle\pm.012}$ & $0.665{\scriptstyle\pm.005}$ & $0.665{\scriptstyle\pm.006}$ & $\mathbf{0.687}{\scriptstyle\pm.002}$ & $0.677{\scriptstyle\pm.004}$ \\
FB v4 & $0.676{\scriptstyle\pm.007}$ & $0.677{\scriptstyle\pm.002}$ & $0.682{\scriptstyle\pm.003}$ & $\mathbf{0.701}{\scriptstyle\pm.002}$ & $0.690{\scriptstyle\pm.001}$ \\
WN v1 & $0.775{\scriptstyle\pm.010}$ & $0.774{\scriptstyle\pm.006}$ & $\mathbf{0.800}{\scriptstyle\pm.005}$ & $0.793{\scriptstyle\pm.008}$ & $\mathbf{0.800}{\scriptstyle\pm.015}$ \\
WN v2 & $0.770{\scriptstyle\pm.005}$ & $0.770{\scriptstyle\pm.008}$ & $0.782{\scriptstyle\pm.003}$ & $0.780{\scriptstyle\pm.008}$ & $\mathbf{0.791}{\scriptstyle\pm.006}$ \\
WN v3 & $0.499{\scriptstyle\pm.011}$ & $0.548{\scriptstyle\pm.009}$ & $0.550{\scriptstyle\pm.008}$ & $\mathbf{0.604}{\scriptstyle\pm.015}$ & $0.600{\scriptstyle\pm.012}$ \\
WN v4 & $0.708{\scriptstyle\pm.004}$ & $0.707{\scriptstyle\pm.011}$ & $0.723{\scriptstyle\pm.002}$ & $0.727{\scriptstyle\pm.004}$ & $\mathbf{0.732}{\scriptstyle\pm.004}$ \\
NELL v1 & $0.785{\scriptstyle\pm.053}$ & $0.833{\scriptstyle\pm.108}$ & $\mathbf{0.904}{\scriptstyle\pm.028}$ & $0.871{\scriptstyle\pm.022}$ & $0.864{\scriptstyle\pm.050}$ \\
NELL v2 & $0.729{\scriptstyle\pm.013}$ & $0.756{\scriptstyle\pm.015}$ & $0.748{\scriptstyle\pm.013}$ & $\mathbf{0.777}{\scriptstyle\pm.020}$ & $0.765{\scriptstyle\pm.010}$ \\
NELL v3 & $0.703{\scriptstyle\pm.027}$ & $0.697{\scriptstyle\pm.031}$ & $0.732{\scriptstyle\pm.012}$ & $\mathbf{0.777}{\scriptstyle\pm.016}$ & $0.766{\scriptstyle\pm.008}$ \\
NELL v4 & $0.730{\scriptstyle\pm.018}$ & $0.722{\scriptstyle\pm.010}$ & $0.758{\scriptstyle\pm.012}$ & $\mathbf{0.777}{\scriptstyle\pm.008}$ & $0.769{\scriptstyle\pm.005}$ \\
ILPC Small & $0.444{\scriptstyle\pm.006}$ & $0.448{\scriptstyle\pm.003}$ & $0.452{\scriptstyle\pm.003}$ & $\mathbf{0.463}{\scriptstyle\pm.002}$ & $0.455{\scriptstyle\pm.003}$ \\
ILPC Large & $0.420{\scriptstyle\pm.004}$ & $0.421{\scriptstyle\pm.006}$ & $0.424{\scriptstyle\pm.004}$ & $\mathbf{0.439}{\scriptstyle\pm.001}$ & $0.429{\scriptstyle\pm.003}$ \\
HM 1k & $0.128{\scriptstyle\pm.009}$ & $0.123{\scriptstyle\pm.023}$ & $0.120{\scriptstyle\pm.018}$ & $0.123{\scriptstyle\pm.008}$ & $\mathbf{0.134}{\scriptstyle\pm.011}$ \\
HM 3k & $0.114{\scriptstyle\pm.007}$ & $0.100{\scriptstyle\pm.014}$ & $0.109{\scriptstyle\pm.011}$ & $0.122{\scriptstyle\pm.006}$ & $\mathbf{0.128}{\scriptstyle\pm.007}$ \\
HM 5k & $0.103{\scriptstyle\pm.007}$ & $0.088{\scriptstyle\pm.013}$ & $0.101{\scriptstyle\pm.010}$ & $\mathbf{0.124}{\scriptstyle\pm.008}$ & $0.118{\scriptstyle\pm.006}$ \\
IndigoBM & $0.646{\scriptstyle\pm.006}$ & $0.639{\scriptstyle\pm.002}$ & $0.643{\scriptstyle\pm.008}$ & $0.635{\scriptstyle\pm.003}$ & $\mathbf{0.653}{\scriptstyle\pm.002}$ \\
\midrule
\multicolumn{6}{@{}l}{\textit{\famthree (23)}} \\
\midrule
FB-25 & $0.642{\scriptstyle\pm.007}$ & $0.636{\scriptstyle\pm.003}$ & $0.643{\scriptstyle\pm.005}$ & $\mathbf{0.663}{\scriptstyle\pm.000}$ & $0.648{\scriptstyle\pm.007}$ \\
FB-50 & $0.547{\scriptstyle\pm.005}$ & $0.544{\scriptstyle\pm.003}$ & $0.528{\scriptstyle\pm.013}$ & $\mathbf{0.564}{\scriptstyle\pm.004}$ & $0.547{\scriptstyle\pm.008}$ \\
FB-75 & $0.607{\scriptstyle\pm.006}$ & $0.603{\scriptstyle\pm.009}$ & $0.597{\scriptstyle\pm.010}$ & $\mathbf{0.624}{\scriptstyle\pm.001}$ & $0.601{\scriptstyle\pm.009}$ \\
FB-100 & $0.633{\scriptstyle\pm.006}$ & $0.618{\scriptstyle\pm.009}$ & $0.633{\scriptstyle\pm.006}$ & $\mathbf{0.660}{\scriptstyle\pm.001}$ & $0.628{\scriptstyle\pm.005}$ \\
WK-25 & $0.502{\scriptstyle\pm.013}$ & $\mathbf{0.504}{\scriptstyle\pm.011}$ & $0.493{\scriptstyle\pm.008}$ & $0.480{\scriptstyle\pm.012}$ & $0.502{\scriptstyle\pm.016}$ \\
WK-50 & $0.301{\scriptstyle\pm.010}$ & $\mathbf{0.314}{\scriptstyle\pm.002}$ & $0.307{\scriptstyle\pm.006}$ & $0.271{\scriptstyle\pm.018}$ & $0.313{\scriptstyle\pm.007}$ \\
WK-75 & $\mathbf{0.542}{\scriptstyle\pm.007}$ & $0.529{\scriptstyle\pm.018}$ & $0.511{\scriptstyle\pm.004}$ & $0.539{\scriptstyle\pm.007}$ & $0.531{\scriptstyle\pm.016}$ \\
WK-100 & $0.288{\scriptstyle\pm.010}$ & $0.288{\scriptstyle\pm.007}$ & $0.292{\scriptstyle\pm.005}$ & $\mathbf{0.301}{\scriptstyle\pm.007}$ & $0.298{\scriptstyle\pm.005}$ \\
NL-0 & $0.547{\scriptstyle\pm.028}$ & $0.511{\scriptstyle\pm.013}$ & $0.539{\scriptstyle\pm.008}$ & $\mathbf{0.594}{\scriptstyle\pm.010}$ & $0.580{\scriptstyle\pm.012}$ \\
NL-25 & $0.544{\scriptstyle\pm.030}$ & $0.509{\scriptstyle\pm.010}$ & $\mathbf{0.582}{\scriptstyle\pm.009}$ & $0.564{\scriptstyle\pm.031}$ & $0.568{\scriptstyle\pm.041}$ \\
NL-50 & $\mathbf{0.552}{\scriptstyle\pm.006}$ & $0.523{\scriptstyle\pm.018}$ & $0.548{\scriptstyle\pm.007}$ & $0.537{\scriptstyle\pm.030}$ & $0.549{\scriptstyle\pm.032}$ \\
NL-75 & $\mathbf{0.521}{\scriptstyle\pm.019}$ & $0.509{\scriptstyle\pm.013}$ & $0.520{\scriptstyle\pm.005}$ & $0.504{\scriptstyle\pm.033}$ & $0.504{\scriptstyle\pm.041}$ \\
NL-100 & $0.655{\scriptstyle\pm.017}$ & $0.650{\scriptstyle\pm.003}$ & $0.670{\scriptstyle\pm.010}$ & $\mathbf{0.673}{\scriptstyle\pm.024}$ & $0.666{\scriptstyle\pm.028}$ \\
Metafam & $0.809{\scriptstyle\pm.066}$ & $0.792{\scriptstyle\pm.060}$ & $0.826{\scriptstyle\pm.053}$ & $0.928{\scriptstyle\pm.004}$ & $\mathbf{0.963}{\scriptstyle\pm.024}$ \\
FBNELL & $0.650{\scriptstyle\pm.016}$ & $0.646{\scriptstyle\pm.026}$ & $0.655{\scriptstyle\pm.013}$ & $\mathbf{0.697}{\scriptstyle\pm.026}$ & $0.674{\scriptstyle\pm.010}$ \\
Wiki MT1 tax & $0.304{\scriptstyle\pm.010}$ & $0.426{\scriptstyle\pm.047}$ & $\mathbf{0.453}{\scriptstyle\pm.015}$ & $0.377{\scriptstyle\pm.016}$ & $0.447{\scriptstyle\pm.061}$ \\
Wiki MT1 health & $0.397{\scriptstyle\pm.037}$ & $0.421{\scriptstyle\pm.023}$ & $0.456{\scriptstyle\pm.003}$ & $\mathbf{0.475}{\scriptstyle\pm.003}$ & $0.461{\scriptstyle\pm.008}$ \\
Wiki MT2 org & $0.148{\scriptstyle\pm.006}$ & $0.142{\scriptstyle\pm.012}$ & $0.154{\scriptstyle\pm.004}$ & $\mathbf{0.165}{\scriptstyle\pm.002}$ & $0.155{\scriptstyle\pm.006}$ \\
Wiki MT2 sci & $0.395{\scriptstyle\pm.046}$ & $0.434{\scriptstyle\pm.052}$ & $0.464{\scriptstyle\pm.016}$ & $0.443{\scriptstyle\pm.005}$ & $\mathbf{0.474}{\scriptstyle\pm.047}$ \\
Wiki MT3 art & $0.418{\scriptstyle\pm.007}$ & $0.418{\scriptstyle\pm.007}$ & $0.447{\scriptstyle\pm.008}$ & $\mathbf{0.469}{\scriptstyle\pm.001}$ & $0.444{\scriptstyle\pm.007}$ \\
Wiki MT3 infra & $0.763{\scriptstyle\pm.021}$ & $0.772{\scriptstyle\pm.013}$ & $0.785{\scriptstyle\pm.010}$ & $\mathbf{0.818}{\scriptstyle\pm.004}$ & $0.775{\scriptstyle\pm.016}$ \\
Wiki MT4 sci & $0.448{\scriptstyle\pm.007}$ & $0.439{\scriptstyle\pm.012}$ & $0.452{\scriptstyle\pm.005}$ & $\mathbf{0.470}{\scriptstyle\pm.014}$ & $0.454{\scriptstyle\pm.016}$ \\
Wiki MT4 health & $0.734{\scriptstyle\pm.029}$ & $0.769{\scriptstyle\pm.006}$ & $0.775{\scriptstyle\pm.004}$ & $\mathbf{0.780}{\scriptstyle\pm.002}$ & $0.719{\scriptstyle\pm.038}$ \\
\midrule
Average ($57$) & $0.523{\scriptstyle\pm.004}$ & $0.524{\scriptstyle\pm.001}$ & $0.542{\scriptstyle\pm.003}$ & $0.548{\scriptstyle\pm.005}$ & $\mathbf{0.551}{\scriptstyle\pm.001}$ \\
\bottomrule
\end{tabular}

\end{table}

\clearpage
\subsection{Link prediction per \halflink scenario}\label{app:ft}

For a test triple $(h,r,t)$, the query half $(h,r,?)$ is seen iff $\exists\,e\in\Vi\setminus\{t\}:(h,r,e)\in\Ei$, and the answer half $(?,r,t)$ iff $\exists\,e\in\Vi\setminus\{h\}:(e,r,t)\in\Ei$~\citep{DBLP:journals/corr/abs-2606-18001}.
Crossing the two outcomes gives the four scenarios \SQSA (Seen Query, Seen Answer), \SQUA (Seen Query, Unseen Answer), \UQSA (Unseen Query, Seen Answer) and \UQUA (Unseen Query, Unseen Answer).
The $50$ zero-shot benchmarks of \cref{tab:vis50} are the $54$ of \cref{sec:lp_results} without NELL995, NELL v1 and Metafam, on which at least one scenario is empty, and AristoV4, on which \MOTIF has a single run~\citep{DBLP:journals/corr/abs-2606-18001}.
In \cref{tab:ft-family3}, we aggregate the zero-shot and fine-tuned MRR per scenario over $22$ of the $23$ \famthree benchmarks, without Metafam, on which at least one scenario is empty.

\begin{table}[!ht]
\centering\small
\caption{\textbf{Fine-tuning does not lower the MRR of \MERIT on \SQUA and \UQUA, as it does for \TRIX}. Mean MRR over the $22$ \famthree benchmarks of \cref{sec:lp_results}, zero-shot and after per-benchmark fine-tuning, averaged over five pre-training runs per model, three for \Flock and \MOTIF; $\pm$ is the sample standard deviation over runs. Bold as in \cref{tab:family-avg}.}
\label{tab:ft-family3}
\begin{tabular}{@{}l rrrrr@{}}
\toprule
model & \Orig & \SQSA & \UQSA & \SQUA & \UQUA\\
\midrule
\multicolumn{6}{@{}l}{\textit{zero-shot}}\\
\midrule
\ULTRA & $0.342_{\pm0.004}$ & $\mathbf{0.381}_{\pm0.006}$ & $0.543_{\pm0.011}$ & $0.113_{\pm0.007}$ & $\mathbf{0.221}_{\pm0.013}$\\
\MOTIF & $0.345_{\pm0.006}$ & $0.370_{\pm0.003}$ & $0.547_{\pm0.010}$ & $0.132_{\pm0.009}$ & $0.214_{\pm0.017}$\\
\TRIX & $\mathbf{0.364}_{\pm0.003}$ & $0.372_{\pm0.004}$ & $\mathbf{0.573}_{\pm0.006}$ & $\mathbf{0.163}_{\pm0.007}$ & $\mathbf{0.221}_{\pm0.014}$\\
\Flock & $\mathbf{0.359}_{\pm0.008}$ & $\mathbf{0.373}_{\pm0.011}$ & $0.561_{\pm0.009}$ & $0.155_{\pm0.004}$ & $\mathbf{0.231}_{\pm0.017}$\\
\MERIT & $0.358_{\pm0.005}$ & $\mathbf{0.380}_{\pm0.004}$ & $\mathbf{0.568}_{\pm0.014}$ & $0.144_{\pm0.008}$ & $\mathbf{0.233}_{\pm0.011}$\\
\midrule
\multicolumn{6}{@{}l}{\textit{fine-tuned}}\\
\midrule
\ULTRA & $0.375_{\pm0.004}$ & $0.396_{\pm0.002}$ & $0.579_{\pm0.003}$ & $0.155_{\pm0.012}$ & $0.251_{\pm0.002}$\\
\MOTIF & $0.373_{\pm0.003}$ & $0.395_{\pm0.002}$ & $0.585_{\pm0.002}$ & $0.151_{\pm0.008}$ & $0.239_{\pm0.004}$\\
\TRIX & $0.368_{\pm0.003}$ & $0.396_{\pm0.004}$ & $0.593_{\pm0.003}$ & $0.126_{\pm0.008}$ & $0.196_{\pm0.010}$\\
\Flock & $\mathbf{0.391}_{\pm0.001}$ & $\mathbf{0.404}_{\pm0.002}$ & $\mathbf{0.599}_{\pm0.002}$ & $\mathbf{0.177}_{\pm0.002}$ & $\mathbf{0.258}_{\pm0.002}$\\
\MERIT & $0.384_{\pm0.001}$ & $\mathbf{0.402}_{\pm0.002}$ & $0.588_{\pm0.002}$ & $0.167_{\pm0.003}$ & $0.255_{\pm0.002}$\\
\bottomrule
\end{tabular}

\end{table}

\clearpage
\subsection{Fine-tuned link prediction per benchmark}\label{app:ft23}

We follow the fine-tuning protocol of \ULTRA~\citep{DBLP:conf/iclr/0001YM0Z24}: three epochs over the triples of $\Gtr$, keeping the checkpoint with the best validation MRR out of the three.
\cref{tab:ft23-mrr,tab:ft23-hits} give the fine-tuned MRR and Hits@10 on each of the $23$ \famthree benchmarks of \cref{sec:lp_results}.

\begin{table}[!ht]
\centering\scriptsize
\caption{\textbf{Fine-tuned MRR of \MERIT and the SoTA KGFMs on each of the $23$ \famthree benchmarks}. Mean MRR after per-benchmark fine-tuning, averaged over five pre-training runs per model, three for \MOTIF and \Flock; $\pm$ is the sample standard deviation over runs. \textbf{Bold} marks the best mean of each row.}
\label{tab:ft23-mrr}
\begin{tabular}{@{}l ccccc@{}}
\toprule
\textbf{Benchmark} & \textbf{\ULTRA} & \textbf{\MOTIF} & \textbf{\TRIX} & \textbf{\Flock} & \textbf{\MERIT} \\
\midrule
FB-25 & $0.387{\scriptstyle\pm.003}$ & $0.386{\scriptstyle\pm.004}$ & $0.379{\scriptstyle\pm.003}$ & $\mathbf{0.404}{\scriptstyle\pm.000}$ & $0.394{\scriptstyle\pm.004}$ \\
FB-50 & $0.337{\scriptstyle\pm.002}$ & $0.335{\scriptstyle\pm.003}$ & $0.323{\scriptstyle\pm.007}$ & $\mathbf{0.357}{\scriptstyle\pm.004}$ & $0.339{\scriptstyle\pm.002}$ \\
FB-75 & $0.406{\scriptstyle\pm.004}$ & $0.400{\scriptstyle\pm.003}$ & $0.390{\scriptstyle\pm.003}$ & $\mathbf{0.423}{\scriptstyle\pm.002}$ & $0.409{\scriptstyle\pm.004}$ \\
FB-100 & $0.439{\scriptstyle\pm.004}$ & $0.438{\scriptstyle\pm.003}$ & $0.433{\scriptstyle\pm.011}$ & $\mathbf{0.458}{\scriptstyle\pm.001}$ & $0.453{\scriptstyle\pm.003}$ \\
WK-25 & $0.308{\scriptstyle\pm.005}$ & $\mathbf{0.316}{\scriptstyle\pm.004}$ & $0.291{\scriptstyle\pm.002}$ & $0.290{\scriptstyle\pm.006}$ & $0.307{\scriptstyle\pm.004}$ \\
WK-50 & $0.146{\scriptstyle\pm.004}$ & $\mathbf{0.153}{\scriptstyle\pm.011}$ & $0.147{\scriptstyle\pm.008}$ & $0.136{\scriptstyle\pm.002}$ & $0.149{\scriptstyle\pm.010}$ \\
WK-75 & $0.378{\scriptstyle\pm.004}$ & $0.376{\scriptstyle\pm.016}$ & $0.368{\scriptstyle\pm.005}$ & $\mathbf{0.405}{\scriptstyle\pm.002}$ & $0.375{\scriptstyle\pm.004}$ \\
WK-100 & $0.163{\scriptstyle\pm.004}$ & $0.171{\scriptstyle\pm.005}$ & $0.177{\scriptstyle\pm.008}$ & $\mathbf{0.183}{\scriptstyle\pm.002}$ & $0.166{\scriptstyle\pm.007}$ \\
NL-0 & $0.339{\scriptstyle\pm.004}$ & $0.319{\scriptstyle\pm.002}$ & $0.335{\scriptstyle\pm.012}$ & $\mathbf{0.395}{\scriptstyle\pm.012}$ & $0.359{\scriptstyle\pm.014}$ \\
NL-25 & $0.396{\scriptstyle\pm.008}$ & $0.381{\scriptstyle\pm.009}$ & $0.371{\scriptstyle\pm.007}$ & $\mathbf{0.398}{\scriptstyle\pm.002}$ & $0.395{\scriptstyle\pm.008}$ \\
NL-50 & $0.412{\scriptstyle\pm.006}$ & $0.408{\scriptstyle\pm.003}$ & $0.401{\scriptstyle\pm.005}$ & $0.403{\scriptstyle\pm.006}$ & $\mathbf{0.413}{\scriptstyle\pm.003}$ \\
NL-75 & $\mathbf{0.373}{\scriptstyle\pm.004}$ & $0.355{\scriptstyle\pm.007}$ & $0.341{\scriptstyle\pm.002}$ & $0.350{\scriptstyle\pm.013}$ & $0.366{\scriptstyle\pm.004}$ \\
NL-100 & $0.468{\scriptstyle\pm.003}$ & $0.463{\scriptstyle\pm.007}$ & $0.466{\scriptstyle\pm.004}$ & $\mathbf{0.489}{\scriptstyle\pm.004}$ & $0.472{\scriptstyle\pm.006}$ \\
Metafam & $0.998{\scriptstyle\pm.002}$ & $\mathbf{1.000}{\scriptstyle\pm.000}$ & $0.991{\scriptstyle\pm.011}$ & $\mathbf{1.000}{\scriptstyle\pm.000}$ & $\mathbf{1.000}{\scriptstyle\pm.000}$ \\
FBNELL & $0.483{\scriptstyle\pm.014}$ & $0.474{\scriptstyle\pm.016}$ & $0.480{\scriptstyle\pm.004}$ & $\mathbf{0.524}{\scriptstyle\pm.004}$ & $0.502{\scriptstyle\pm.005}$ \\
Wiki MT1 tax & $0.390{\scriptstyle\pm.071}$ & $0.398{\scriptstyle\pm.083}$ & $0.402{\scriptstyle\pm.011}$ & $0.428{\scriptstyle\pm.002}$ & $\mathbf{0.442}{\scriptstyle\pm.007}$ \\
Wiki MT1 health & $0.379{\scriptstyle\pm.005}$ & $0.384{\scriptstyle\pm.000}$ & $0.380{\scriptstyle\pm.001}$ & $\mathbf{0.390}{\scriptstyle\pm.002}$ & $0.385{\scriptstyle\pm.002}$ \\
Wiki MT2 org & $0.103{\scriptstyle\pm.001}$ & $0.105{\scriptstyle\pm.000}$ & $0.099{\scriptstyle\pm.001}$ & $\mathbf{0.106}{\scriptstyle\pm.000}$ & $0.104{\scriptstyle\pm.001}$ \\
Wiki MT2 sci & $0.352{\scriptstyle\pm.027}$ & $0.332{\scriptstyle\pm.002}$ & $0.343{\scriptstyle\pm.006}$ & $\mathbf{0.389}{\scriptstyle\pm.003}$ & $0.365{\scriptstyle\pm.014}$ \\
Wiki MT3 art & $0.314{\scriptstyle\pm.002}$ & $0.317{\scriptstyle\pm.001}$ & $0.277{\scriptstyle\pm.006}$ & $\mathbf{0.329}{\scriptstyle\pm.002}$ & $0.322{\scriptstyle\pm.002}$ \\
Wiki MT3 infra & $0.666{\scriptstyle\pm.004}$ & $0.690{\scriptstyle\pm.002}$ & $0.683{\scriptstyle\pm.006}$ & $\mathbf{0.709}{\scriptstyle\pm.000}$ & $0.697{\scriptstyle\pm.002}$ \\
Wiki MT4 sci & $0.310{\scriptstyle\pm.002}$ & $0.310{\scriptstyle\pm.004}$ & $0.306{\scriptstyle\pm.004}$ & $\mathbf{0.330}{\scriptstyle\pm.001}$ & $0.315{\scriptstyle\pm.001}$ \\
Wiki MT4 health & $0.701{\scriptstyle\pm.003}$ & $0.703{\scriptstyle\pm.002}$ & $0.701{\scriptstyle\pm.003}$ & $0.708{\scriptstyle\pm.003}$ & $\mathbf{0.710}{\scriptstyle\pm.001}$ \\
\midrule
Average ($23$) & $0.402{\scriptstyle\pm.004}$ & $0.401{\scriptstyle\pm.003}$ & $0.395{\scriptstyle\pm.003}$ & $\mathbf{0.418}{\scriptstyle\pm.001}$ & $0.410{\scriptstyle\pm.001}$ \\
\bottomrule
\end{tabular}

\end{table}

\begin{table}[!ht]
\centering\scriptsize
\caption{\textbf{Fine-tuned Hits@10 of \MERIT and the SoTA KGFMs on each of the $23$ \famthree benchmarks}. Mean Hits@10 after per-benchmark fine-tuning, averaged over five pre-training runs per model, three for \MOTIF and \Flock; $\pm$ is the sample standard deviation over runs. \textbf{Bold} marks the best mean of each row.}
\label{tab:ft23-hits}
\begin{tabular}{@{}l ccccc@{}}
\toprule
\textbf{Benchmark} & \textbf{\ULTRA} & \textbf{\MOTIF} & \textbf{\TRIX} & \textbf{\Flock} & \textbf{\MERIT} \\
\midrule
FB-25 & $0.640{\scriptstyle\pm.005}$ & $0.633{\scriptstyle\pm.006}$ & $0.629{\scriptstyle\pm.006}$ & $\mathbf{0.664}{\scriptstyle\pm.001}$ & $0.645{\scriptstyle\pm.004}$ \\
FB-50 & $0.543{\scriptstyle\pm.003}$ & $0.538{\scriptstyle\pm.007}$ & $0.526{\scriptstyle\pm.008}$ & $\mathbf{0.570}{\scriptstyle\pm.002}$ & $0.547{\scriptstyle\pm.005}$ \\
FB-75 & $0.610{\scriptstyle\pm.002}$ & $0.599{\scriptstyle\pm.007}$ & $0.586{\scriptstyle\pm.006}$ & $\mathbf{0.631}{\scriptstyle\pm.001}$ & $0.611{\scriptstyle\pm.004}$ \\
FB-100 & $0.642{\scriptstyle\pm.005}$ & $0.635{\scriptstyle\pm.006}$ & $0.628{\scriptstyle\pm.010}$ & $\mathbf{0.666}{\scriptstyle\pm.006}$ & $0.648{\scriptstyle\pm.004}$ \\
WK-25 & $0.497{\scriptstyle\pm.010}$ & $0.505{\scriptstyle\pm.004}$ & $0.491{\scriptstyle\pm.009}$ & $0.465{\scriptstyle\pm.013}$ & $\mathbf{0.513}{\scriptstyle\pm.005}$ \\
WK-50 & $\mathbf{0.298}{\scriptstyle\pm.003}$ & $0.289{\scriptstyle\pm.007}$ & $0.288{\scriptstyle\pm.010}$ & $0.270{\scriptstyle\pm.022}$ & $0.290{\scriptstyle\pm.018}$ \\
WK-75 & $0.522{\scriptstyle\pm.014}$ & $0.535{\scriptstyle\pm.011}$ & $0.512{\scriptstyle\pm.008}$ & $\mathbf{0.558}{\scriptstyle\pm.002}$ & $0.530{\scriptstyle\pm.008}$ \\
WK-100 & $0.277{\scriptstyle\pm.003}$ & $0.286{\scriptstyle\pm.004}$ & $0.290{\scriptstyle\pm.005}$ & $\mathbf{0.300}{\scriptstyle\pm.006}$ & $0.286{\scriptstyle\pm.008}$ \\
NL-0 & $0.555{\scriptstyle\pm.023}$ & $0.542{\scriptstyle\pm.004}$ & $0.516{\scriptstyle\pm.013}$ & $\mathbf{0.626}{\scriptstyle\pm.014}$ & $0.583{\scriptstyle\pm.030}$ \\
NL-25 & $0.606{\scriptstyle\pm.012}$ & $0.567{\scriptstyle\pm.011}$ & $0.528{\scriptstyle\pm.019}$ & $\mathbf{0.615}{\scriptstyle\pm.011}$ & $0.602{\scriptstyle\pm.021}$ \\
NL-50 & $\mathbf{0.593}{\scriptstyle\pm.006}$ & $0.574{\scriptstyle\pm.011}$ & $0.549{\scriptstyle\pm.006}$ & $0.591{\scriptstyle\pm.010}$ & $0.588{\scriptstyle\pm.006}$ \\
NL-75 & $\mathbf{0.574}{\scriptstyle\pm.002}$ & $0.552{\scriptstyle\pm.003}$ & $0.512{\scriptstyle\pm.008}$ & $0.555{\scriptstyle\pm.011}$ & $0.566{\scriptstyle\pm.006}$ \\
NL-100 & $0.678{\scriptstyle\pm.006}$ & $0.677{\scriptstyle\pm.007}$ & $0.672{\scriptstyle\pm.012}$ & $\mathbf{0.718}{\scriptstyle\pm.002}$ & $0.691{\scriptstyle\pm.009}$ \\
Metafam & $\mathbf{1.000}{\scriptstyle\pm.000}$ & $\mathbf{1.000}{\scriptstyle\pm.000}$ & $\mathbf{1.000}{\scriptstyle\pm.000}$ & $\mathbf{1.000}{\scriptstyle\pm.000}$ & $\mathbf{1.000}{\scriptstyle\pm.000}$ \\
FBNELL & $0.650{\scriptstyle\pm.011}$ & $0.637{\scriptstyle\pm.010}$ & $0.641{\scriptstyle\pm.010}$ & $\mathbf{0.709}{\scriptstyle\pm.007}$ & $0.676{\scriptstyle\pm.012}$ \\
Wiki MT1 tax & $0.498{\scriptstyle\pm.073}$ & $0.516{\scriptstyle\pm.027}$ & $0.518{\scriptstyle\pm.001}$ & $0.508{\scriptstyle\pm.003}$ & $\mathbf{0.537}{\scriptstyle\pm.002}$ \\
Wiki MT1 health & $0.466{\scriptstyle\pm.010}$ & $0.469{\scriptstyle\pm.002}$ & $0.474{\scriptstyle\pm.004}$ & $\mathbf{0.489}{\scriptstyle\pm.001}$ & $0.473{\scriptstyle\pm.005}$ \\
Wiki MT2 org & $0.169{\scriptstyle\pm.000}$ & $0.173{\scriptstyle\pm.003}$ & $0.163{\scriptstyle\pm.002}$ & $\mathbf{0.176}{\scriptstyle\pm.002}$ & $0.175{\scriptstyle\pm.002}$ \\
Wiki MT2 sci & $\mathbf{0.544}{\scriptstyle\pm.002}$ & $0.528{\scriptstyle\pm.009}$ & $0.532{\scriptstyle\pm.006}$ & $0.537{\scriptstyle\pm.006}$ & $\mathbf{0.544}{\scriptstyle\pm.011}$ \\
Wiki MT3 art & $0.472{\scriptstyle\pm.004}$ & $0.468{\scriptstyle\pm.004}$ & $0.410{\scriptstyle\pm.012}$ & $\mathbf{0.488}{\scriptstyle\pm.003}$ & $0.478{\scriptstyle\pm.004}$ \\
Wiki MT3 infra & $0.810{\scriptstyle\pm.003}$ & $0.827{\scriptstyle\pm.002}$ & $0.820{\scriptstyle\pm.005}$ & $\mathbf{0.840}{\scriptstyle\pm.001}$ & $0.828{\scriptstyle\pm.002}$ \\
Wiki MT4 sci & $0.486{\scriptstyle\pm.003}$ & $0.483{\scriptstyle\pm.008}$ & $0.475{\scriptstyle\pm.002}$ & $\mathbf{0.509}{\scriptstyle\pm.003}$ & $0.485{\scriptstyle\pm.003}$ \\
Wiki MT4 health & $0.784{\scriptstyle\pm.003}$ & $0.786{\scriptstyle\pm.002}$ & $0.783{\scriptstyle\pm.002}$ & $\mathbf{0.790}{\scriptstyle\pm.000}$ & $0.788{\scriptstyle\pm.002}$ \\
\midrule
Average ($23$) & $0.561{\scriptstyle\pm.004}$ & $0.557{\scriptstyle\pm.002}$ & $0.545{\scriptstyle\pm.004}$ & $\mathbf{0.577}{\scriptstyle\pm.001}$ & $0.569{\scriptstyle\pm.001}$ \\
\bottomrule
\end{tabular}

\end{table}

\subsection{Ablations of \MERIT}\label{app:ablations}

\cref{tab:ablations} gives the MRR of every variant of \MERIT in \cref{sec:lp_results}. With the \textit{motif bias only}, the tokens carry no relation signature, and with the \textit{relation signature only}, the attention logits carry no motif bias.
Every variant keeps the noise and the query flag of \cref{eq:init}, and is pre-trained as \MERIT is (\cref{app:impl}).

\begin{table}[!ht]
\centering\small
\caption{\textbf{Both the attention over all relation names and the motifs as a soft prior contribute to the MRR of \MERIT}. Mean MRR over the $54$ benchmarks of \cref{sec:lp_results}, averaged over the pre-training runs in the second column; $\pm$ is the sample standard deviation over runs.}
\label{tab:ablations}
\begin{tabular}{@{}l c c@{}}
\toprule
\textbf{Model} & \textbf{runs} & \textbf{MRR} \\
\midrule
\MERIT (motif bias and relation signature) & 5 & $0.386{\scriptstyle\pm.002}$ \\
\midrule
motif bias only & 3 & $0.379{\scriptstyle\pm.002}$ \\
relation signature only & 1 & $0.292{\scriptstyle\pm.000}$ \\
neither (plain transformer) & 1 & $0.312{\scriptstyle\pm.000}$ \\
\midrule
\ULTRA & 5 & $0.365{\scriptstyle\pm.003}$ \\
\midrule
masked \MERIT & 3 & $0.368{\scriptstyle\pm.007}$ \\
\bottomrule
\end{tabular}

\end{table}

\clearpage
\section{Complex query answering}\label{app:cqa}

\subsection{Complex queries and query types}\label{app:cqa-queries}

\begin{figure}[t]
\renewcommand{\sfdefault}{cmss}
\renewcommand{\bfdefault}{bx}
\centering

\tikzset{
  anchQT/.style = {draw=petroil2, circle, inner sep=0pt, minimum size=10pt, line width=2pt},
  varQT/.style  = {draw=gold2, circle, inner sep=0pt, minimum size=10pt, line width=2pt},
  tgtQT/.style  = {draw=lacamlightlilac, circle, inner sep=0pt, minimum size=10pt, line width=2pt},
  conjQT/.style = {-{Stealth[length=5pt]}, bgrey2, line width=1.2pt, shorten >=0.5pt},
  disjQT/.style = {-{Diamond[length=4.6pt 1.1, width=3.8pt 1.1]}, olive3, line width=1.2pt, shorten >=0.5pt},
  nconjQT/.style= {-{Stealth[length=5pt]}, pink2, line width=1.2pt, shorten >=0.5pt},
  ndisjQT/.style= {-{Diamond[length=4.6pt 1.1, width=3.8pt 1.1]}, pink2, line width=1.2pt, shorten >=0.5pt},
  badgeQT/.style= {draw=tomato0!70!tomato1, fill=tomato0!70!tomato1, inner sep=2.5pt,
                   rounded corners=1.5pt, font=\scriptsize\bfseries\sffamily, text=white},
  grpQT/.style  = {text=bgrey4, font=\scriptsize\sffamily, inner sep=1pt, anchor=east},
  keyQT/.style  = {text=bgrey4, font=\scriptsize\sffamily, inner sep=1pt, anchor=west},
  panelQT/.style= {},
}

\newcommand{\panelQTbox}{\path[use as bounding box] (-13pt,-37pt) rectangle (53pt,8pt);}
\newcommand{\badgeQTat}[2]{\node [badgeQT] at (#1pt,-31pt) {#2};}

\begin{tabular}{@{}r@{\hspace{5pt}}c@{\hspace{1pt}}c@{\hspace{1pt}}c@{\hspace{1pt}}c@{\hspace{1pt}}c@{}}

 &
\includegraphics{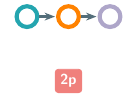} &
\includegraphics{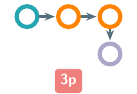} &
\includegraphics{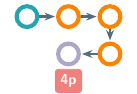} & & \\[2pt]

\raisebox{12pt}{\includegraphics{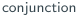}} &
\includegraphics{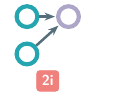} &
\includegraphics{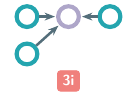} &
\includegraphics{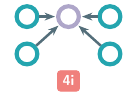} & & \\[2pt]

 &
\includegraphics{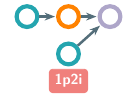} &
\includegraphics{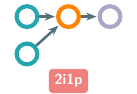} & & & \\[2pt]

\raisebox{12pt}{\includegraphics{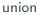}} &
\includegraphics{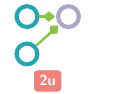} &
\includegraphics{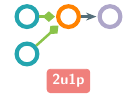} & & & \\[2pt]

\raisebox{12pt}{\includegraphics{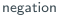}} &
\includegraphics{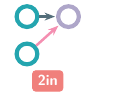} &
\includegraphics{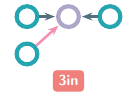} &
\includegraphics{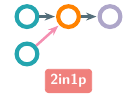} &
\includegraphics{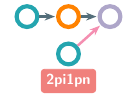} &
\includegraphics{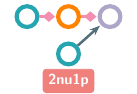} \\

\end{tabular}

\vspace{4pt}

\includegraphics{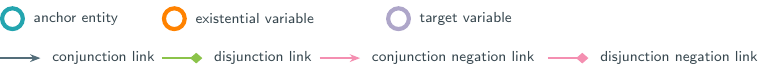}

\caption{\textbf{The $15$ complex query types on which we evaluate CQA}, grouped as in \cref{tab:cqa,tab:cqa-queries}. Structures adapted from \citet{DBLP:conf/nips/RenL20}, with the 4p and 4i types of \citet{DBLP:conf/icml/GregucciX0LMSV25}.}
\label{fig:query-types}
\end{figure}

\cref{fig:query-types} draws the $15$ complex query types on which we evaluate CQA, and \cref{tab:cqa-queries} counts the queries of each type in every benchmark.
Following \cref{sec:kgfms}, $a_i$ are anchor entities, $r_i$ relation names, $y_i$ existentially quantified variables, to be grounded into entities, and $?x$ the target variable.
The path types chain one to four links,
\begin{align}
\tag{1p} &?x : (a_1, r_1, x),\\
\tag{2p} &?x : \exists\, y_1 .\ (a_1, r_1, y_1) \wedge (y_1, r_2, x),
\end{align}
where 3p and 4p extend the chain with $y_2$ and $y_3$.
The intersection types take the intersection of the target entity sets defined over two to four links,
\begin{equation}
\tag{2i} ?x : (a_1, r_1, x) \wedge (a_2, r_2, x),
\end{equation}
where 3i and 4i add a third and a fourth link.
Combining a path and an intersection gives
\begin{equation}
\tag{2i1p} ?x : \exists\, y_1 .\ (a_1, r_1, y_1) \wedge (a_2, r_2, y_1) \wedge (y_1, r_3, x),
\end{equation}
and 1p2i inverts the order of operations.
The union types replace a conjunction with a disjunction,
\begin{equation}
\tag{2u} ?x : (a_1, r_1, x) \vee (a_2, r_2, x),
\end{equation}
and 2u1p follows the disjunction with a link.
The negation types negate a triple pattern of a query type above,
\begin{equation}
\tag{2in} ?x : (a_1, r_1, x) \wedge \neg (a_2, r_2, x),
\end{equation}
where 3in, 2in1p and 2pi1pn negate one triple pattern of 3i, 2i1p and 1p2i, and 2nu1p negates the whole 2p path of 1p2i, $\neg((a_1, r_1, y_1) \wedge (y_1, r_2, x))$, which De Morgan laws rewrite as $\neg(a_1, r_1, y_1) \vee \neg(y_1, r_2, x)$.
In \cref{fig:query-types,tab:cqa,tab:cqa-queries}, the conjunctive types of \cref{sec:cqa_results} are the path types, the intersection types, 1p2i and 2i1p; the union types are 2u and 2u1p; and the negation types are 2in, 3in, 2in1p, 2pi1pn and 2nu1p.

We name the query types as \citet{DBLP:conf/icml/GregucciX0LMSV25} do, writing the number of steps of each path and of each conjunction.

\subsection{Related work on CQA}\label{app:cqa-sota}

As in link prediction (\cref{sec:kgfms}), CQA methods such as CQD~\citep{DBLP:conf/iclr/ArakelyanDMC21}, QTO~\citep{DBLP:conf/icml/BaiLLH23} and FIT~\citep{DBLP:conf/iclr/Yin0S24} rely on link predictors that learn an embedding for every entity and relation name.
GNN-QE~\citep{DBLP:conf/icml/Zhu0Z022} generalizes to unseen entities, but still learns embeddings for relation names.
Hence, none of these methods can answer queries in the inductive transfer setting.

\clearpage
\subsection{The +H benchmarks of \famtwo and \famthree}\label{app:plush}

\citet{DBLP:conf/icml/GregucciX0LMSV25} call a QA pair \textit{full-inference} if its reasoning tree contains only links of the test graph, and \textit{partial-inference} if it contains at least one link of the training graph and at least one of the test graph, in which case the pair reduces to a simpler query type.
Their +H benchmarks balance QA pairs: for each query type, every reduction, that is each simpler query type it reduces to and full-inference, holds $10{,}000$ QA pairs, and no anchor entity and no relation name appears in more than $20\%$ of the QA pairs of a query type.
Following this construction, we rebuild the \famtwo and \famthree benchmarks of \UltraQuery~\citep{DBLP:conf/nips/GalkinZR022,DBLP:conf/nips/GalkinZ00Z24} as +H benchmarks, where the inference graph $\Gi$ takes the place of the training graph and the test graph $\Gu$ holds the links to predict, and keep the +H benchmarks of \famone as released.

Similarly to \citet{DBLP:conf/icml/GregucciX0LMSV25}, we sample queries for one reduction at a time and assign every QA pair to the reduction it falls in, drawing the links of the reasoning tree from $\Gi$ or from $\Gu$ as the reduction prescribes.
Then, as the inference graphs of \famthree carry far fewer relation names than the graphs of \famone, their most frequent relation name covers $11.7\%$ (Science) to $45.9\%$ (Location) of the links, against $5.6\%$ (FB15k237) to $15.9\%$ (ICEWS18).
A cap of $20\%$ would leave few QA pairs to the query types with such a relation name, so we raise it to $25\%$ of the QA pairs of a reduction, on both families.
We fill every reduction of a query type with the same number of QA pairs, the largest number at most $10{,}000$ that every reduction of the type reaches under this cap; it goes from $10{,}000$ down to $4$ for the 4p and 4i queries of Location.
We keep a graph only if every query type has QA pairs under this cap, which excludes the Award, Education, Organization and Sport graphs of \UltraQuery from \famthree.
The 1p queries are the 1p test queries of \UltraQuery, unchanged, as in the +H benchmarks of \famone~\citep{DBLP:conf/icml/GregucciX0LMSV25}.

\subsection{Experimental details}\label{app:cqa-details}

We fine-tune \UltraQuery and \MERITQuery with the configuration that \citet{DBLP:conf/nips/GalkinZ00Z24} release for \UltraQuery, and change only the starting checkpoints: the pre-training runs of \ULTRA and of \MERIT on the three graphs of \cref{sec:lp_results}, in place of the checkpoint pre-trained on four graphs~\citep{DBLP:conf/iclr/0001YM0Z24} from which \citet{DBLP:conf/nips/GalkinZ00Z24} start, so that CQA starts from the same five pre-training runs per model as link prediction. Each fine-tuning run trains on the training queries of \fbnew\ (\cref{tab:cqa-queries}) for ten epochs of $4{,}000$ steps, $40{,}000$ steps in total, with Adam at a learning rate of $5 \times 10^{-4}$ and batches of $32$ queries on each of four GPUs. We answer the queries of all $19$ benchmarks with the fuzzy logic operators of \UltraQuery.

\subsection{Results per query type}\label{app:cqa-types-sec}

\begin{figure}[!ht]
\centering
{\figfonts
\colorlet{mUQct}{petroil4}
\colorlet{mMQct}{olive5}
\pgfplotsset{
  ctaxisV2/.style={
    ybar, bar width=4.2pt, scale only axis, width=12.6cm, height=2.2cm,
    xmin=0.4, xmax=16.6, ymin=0, enlarge y limits={upper, value=0.1},
    xtick={1,...,16}, xticklabels/.expand once={\cqaTypesTicks},
    ytick distance=0.2, yticklabel={\pgfmathprintnumber[fixed, precision=1, assume math mode=true]{\tick}},
    axis lines=left, axis line style={-, bgrey1, line width=0.5pt},
    tick style={bgrey1, line width=0.5pt}, major x tick style={draw=none},
    ymajorgrids, grid style={bgrey0!22, line width=0.4pt},
    tick label style={font=\tiny\sffamily, text=bgrey4},
    ylabel style={font=\scriptsize\sffamily, text=bgrey4},
    title style={font=\scriptsize\sffamily, text=bgrey4, yshift=-2pt},
    error bars/y dir=both, error bars/y explicit,
    error bars/error bar style={bgrey4, line width=0.4pt}, error bars/error mark options={rotate=90, mark size=1pt, bgrey4},
    legend style={font=\scriptsize\sffamily, draw=none, fill=none, legend columns=2, /tikz/every even column/.append style={column sep=8pt}},
    legend image code/.code={\draw[#1, draw=none] (0cm,-0.08cm) rectangle (0.25cm,0.1cm);},
  },
  ctUQV2/.style={fill=mUQct, draw=none},
  ctMQV2/.style={fill=mMQct, draw=none},
}
\includegraphics{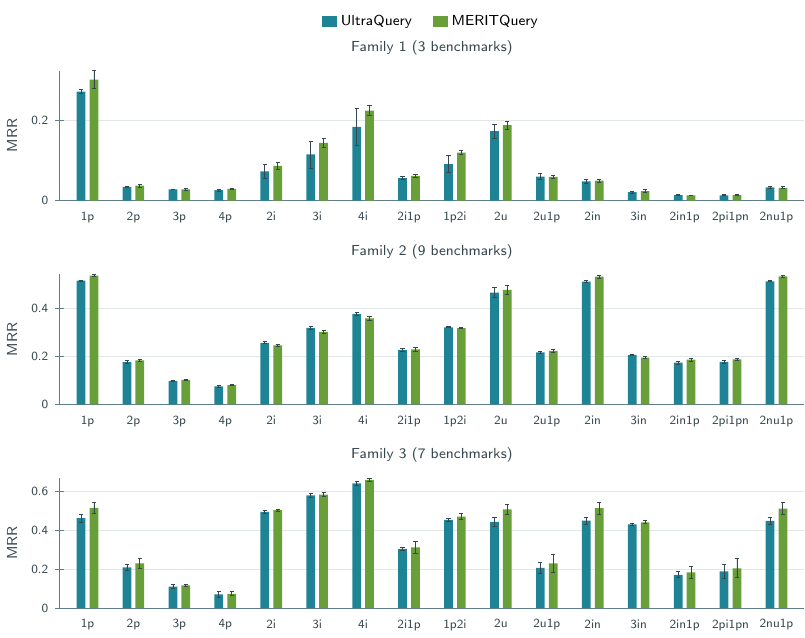}}
\caption{\textbf{\MERITQuery is ahead of \UltraQuery on every query type of \famthree, and on $12$ and $11$ of the $16$ of \famone and \famtwo}. Mean MRR over the benchmarks of each family per query type, averaged over the five fine-tuning runs per method of \cref{sec:cqa_results}, with FB15k237+H in \famone; error bars are the sample standard deviation over runs. 1p appears only here.}
\label{fig:cqa-types}
\end{figure}

\cref{fig:cqa-types} breaks down \cref{tab:cqa} by query type in each family, and includes the 1p queries and \fbnewH.

\clearpage
\section{Benchmark statistics}\label{app:stats}\label{app:cqa-stats-sec}

\cref{tab:lp-stats} lists by family the $57$ link-prediction benchmarks, the $54$ of \cref{sec:lp_results} and the three pre-training graphs; \cref{tab:cqa-stats} lists the graphs behind the $19$ CQA benchmarks, the $18$ of \cref{sec:cqa_results} and \fbnewH; and \cref{tab:cqa-queries} counts the queries of each query type in each of them. All counts of relation names and triples are those of the original graphs: every model adds to $\Gi$ the inverse of each triple (\cref{sec:kgfms}), and so works with twice as many relation names and triples.

\begin{table}[!ht]
\centering\scriptsize
\caption{\textbf{The $57$ link-prediction benchmarks}. Entities, relation names and triples of the inference graph, and the number of test triples, as reported by \citet{DBLP:conf/iclr/0001YM0Z24}.}
\label{tab:lp-stats}
\begin{tabular}{@{}l rrrr@{}}
\toprule
\textbf{Benchmark} & \textbf{Entities} & \textbf{Relation names} & \textbf{Triples} & \textbf{Test triples} \\
\midrule
\multicolumn{5}{@{}l}{\textit{\famone (16)}} \\
\midrule
CoDEx Small & $2{,}034$ & $42$ & $32{,}888$ & $1{,}828$ \\
CoDEx Medium & $17{,}050$ & $51$ & $185{,}584$ & $10{,}311$ \\
CoDEx Large & $77{,}951$ & $69$ & $551{,}193$ & $30{,}622$ \\
WDsinger & $10{,}282$ & $135$ & $16{,}142$ & $2{,}203$ \\
FB15k237\_10 & $11{,}512$ & $237$ & $27{,}211$ & $18{,}150$ \\
FB15k237\_20 & $13{,}166$ & $237$ & $54{,}423$ & $19{,}776$ \\
FB15k237\_50 & $14{,}149$ & $237$ & $136{,}057$ & $20{,}324$ \\
FB15k237 & $14{,}541$ & $237$ & $272{,}115$ & $20{,}466$ \\
NELL23k & $22{,}925$ & $200$ & $25{,}445$ & $4{,}952$ \\
WN18RR & $40{,}943$ & $11$ & $86{,}835$ & $3{,}134$ \\
AristoV4 & $44{,}949$ & $1{,}605$ & $242{,}567$ & $20{,}000$ \\
Hetionet & $45{,}158$ & $24$ & $2{,}025{,}177$ & $112{,}510$ \\
NELL995 & $74{,}536$ & $200$ & $149{,}678$ & $2{,}818$ \\
ConceptNet100k & $78{,}334$ & $34$ & $100{,}000$ & $1{,}200$ \\
DBpedia100k & $99{,}604$ & $470$ & $597{,}572$ & $50{,}000$ \\
YAGO310 & $123{,}182$ & $37$ & $1{,}079{,}040$ & $5{,}000$ \\
\midrule
\multicolumn{5}{@{}l}{\textit{\famtwo (18)}} \\
\midrule
FB v1 & $1{,}093$ & $180$ & $1{,}993$ & $411$ \\
FB v2 & $1{,}660$ & $200$ & $4{,}145$ & $947$ \\
FB v3 & $2{,}501$ & $215$ & $7{,}406$ & $1{,}731$ \\
FB v4 & $3{,}051$ & $219$ & $11{,}714$ & $2{,}840$ \\
WN v1 & $922$ & $9$ & $1{,}618$ & $373$ \\
WN v2 & $2{,}757$ & $10$ & $4{,}011$ & $852$ \\
WN v3 & $5{,}084$ & $11$ & $6{,}327$ & $1{,}143$ \\
WN v4 & $7{,}084$ & $9$ & $12{,}334$ & $2{,}823$ \\
NELL v1 & $225$ & $14$ & $833$ & $201$ \\
NELL v2 & $2{,}086$ & $88$ & $4{,}586$ & $935$ \\
NELL v3 & $3{,}566$ & $142$ & $8{,}048$ & $1{,}620$ \\
NELL v4 & $2{,}795$ & $76$ & $7{,}073$ & $1{,}447$ \\
ILPC Small & $6{,}653$ & $48$ & $20{,}960$ & $2{,}902$ \\
ILPC Large & $29{,}246$ & $65$ & $77{,}044$ & $10{,}184$ \\
HM 1k & $9{,}899$ & $11$ & $18{,}638$ & $476$ \\
HM 3k & $19{,}218$ & $11$ & $38{,}285$ & $1{,}349$ \\
HM 5k & $23{,}792$ & $11$ & $48{,}425$ & $2{,}124$ \\
IndigoBM & $14{,}775$ & $229$ & $250{,}195$ & $14{,}904$ \\
\midrule
\multicolumn{5}{@{}l}{\textit{\famthree (23)}} \\
\midrule
FB-25 & $4{,}097$ & $216$ & $17{,}147$ & $5{,}716$ \\
FB-50 & $4{,}445$ & $205$ & $11{,}636$ & $3{,}879$ \\
FB-75 & $2{,}792$ & $186$ & $9{,}316$ & $3{,}106$ \\
FB-100 & $2{,}624$ & $77$ & $6{,}987$ & $2{,}329$ \\
WK-25 & $3{,}228$ & $74$ & $3{,}391$ & $1{,}131$ \\
WK-50 & $9{,}328$ & $93$ & $9{,}672$ & $3{,}225$ \\
WK-75 & $2{,}722$ & $65$ & $3{,}430$ & $1{,}144$ \\
WK-100 & $12{,}136$ & $37$ & $13{,}487$ & $4{,}496$ \\
NL-0 & $2{,}026$ & $112$ & $2{,}287$ & $763$ \\
NL-25 & $2{,}146$ & $120$ & $2{,}230$ & $744$ \\
NL-50 & $2{,}335$ & $119$ & $2{,}576$ & $859$ \\
NL-75 & $1{,}578$ & $116$ & $1{,}818$ & $607$ \\
NL-100 & $1{,}709$ & $53$ & $2{,}378$ & $793$ \\
Metafam & $656$ & $28$ & $7{,}257$ & $184$ \\
FBNELL & $4{,}752$ & $183$ & $10{,}685$ & $597$ \\
Wiki MT1 tax & $10{,}000$ & $9$ & $16{,}526$ & $1{,}834$ \\
Wiki MT1 health & $10{,}000$ & $7$ & $14{,}110$ & $1{,}566$ \\
Wiki MT2 org & $10{,}000$ & $11$ & $21{,}976$ & $2{,}441$ \\
Wiki MT2 sci & $10{,}000$ & $16$ & $14{,}852$ & $1{,}650$ \\
Wiki MT3 art & $10{,}000$ & $45$ & $28{,}023$ & $3{,}113$ \\
Wiki MT3 infra & $10{,}000$ & $27$ & $21{,}646$ & $2{,}405$ \\
Wiki MT4 sci & $10{,}000$ & $42$ & $12{,}516$ & $1{,}388$ \\
Wiki MT4 health & $10{,}000$ & $20$ & $15{,}337$ & $1{,}703$ \\
\bottomrule
\end{tabular}

\end{table}

\begin{table}[!ht]
\centering\scriptsize
\caption{\textbf{The graphs behind the $19$ CQA benchmarks}, those of \citet{DBLP:conf/icml/GregucciX0LMSV25} for \famone and of \UltraQuery~\citep{DBLP:conf/nips/GalkinZ00Z24} for \famtwo and \famthree. Entities and relation names are those of the inference graph.}
\label{tab:cqa-stats}
\begin{tabular}{@{}l rrrrr@{}}
\toprule
\textbf{Dataset} & \textbf{Entities} & \textbf{Relation names} & \textbf{Training triples} & \textbf{Inference triples} & \textbf{Test triples} \\
\midrule
\multicolumn{6}{@{}l}{\textit{\famone}} \\
\midrule
FB15k237+H & $14{,}505$ & $237$ & $272{,}115$ & -- & $20{,}438$ \\
NELL995+H & $63{,}361$ & $200$ & $114{,}213$ & -- & $14{,}267$ \\
ICEWS18+H & $20{,}840$ & $250$ & $213{,}304$ & -- & $24{,}689$ \\
\midrule
\multicolumn{6}{@{}l}{\textit{\famtwo}} \\
\midrule
FB 106\%+H & $13{,}802$ & $233$ & $246{,}713$ & $269{,}449$ & $4{,}012$ \\
FB 113\%+H & $13{,}021$ & $234$ & $200{,}839$ & $243{,}035$ & $7{,}447$ \\
FB 122\%+H & $12{,}314$ & $233$ & $149{,}440$ & $215{,}447$ & $11{,}645$ \\
FB 134\%+H & $11{,}471$ & $233$ & $114{,}365$ & $183{,}906$ & $12{,}265$ \\
FB 150\%+H & $10{,}782$ & $231$ & $81{,}342$ & $165{,}677$ & $14{,}878$ \\
FB 175\%+H & $9{,}781$ & $218$ & $51{,}261$ & $133{,}248$ & $14{,}446$ \\
FB 217\%+H & $9{,}058$ & $223$ & $26{,}228$ & $106{,}194$ & $14{,}089$ \\
FB 300\%+H & $8{,}266$ & $206$ & $12{,}220$ & $93{,}579$ & $14{,}329$ \\
FB 550\%+H & $7{,}275$ & $156$ & $2{,}633$ & $66{,}763$ & $11{,}252$ \\
\midrule
\multicolumn{6}{@{}l}{\textit{\famthree}} \\
\midrule
WikiTopics Art+H & $10{,}000$ & $45$ & $27{,}262$ & $28{,}023$ & $3{,}113$ \\
WikiTopics Health+H & $10{,}000$ & $20$ & $15{,}539$ & $15{,}337$ & $1{,}703$ \\
WikiTopics Infrastructure+H & $10{,}000$ & $27$ & $21{,}990$ & $21{,}646$ & $2{,}405$ \\
WikiTopics Location+H & $10{,}000$ & $34$ & $85{,}063$ & $80{,}269$ & $8{,}917$ \\
WikiTopics People+H & $10{,}000$ & $25$ & $55{,}698$ & $58{,}530$ & $6{,}503$ \\
WikiTopics Science+H & $10{,}000$ & $42$ & $12{,}576$ & $12{,}516$ & $1{,}388$ \\
WikiTopics Taxonomy+H & $10{,}000$ & $31$ & $18{,}921$ & $19{,}416$ & $2{,}157$ \\
\bottomrule
\end{tabular}

\end{table}

\begin{table}[!ht]
\centering\scriptsize
\setlength{\tabcolsep}{2pt}
\caption{\textbf{The queries of the $19$ CQA benchmarks}. Number of test queries per query type, and of training and validation queries of FB15k237, the fine-tuning dataset of \UltraQuery and \MERITQuery; -- marks a query type without queries in the split.}
\label{tab:cqa-queries}
\resizebox{\textwidth}{!}{\begin{tabular}{@{}l@{\hspace{8pt}}rrrrrrrrr@{\hspace{8pt}}rr@{\hspace{8pt}}rrrrr@{}}
\toprule
 & \multicolumn{9}{c}{conjunction} & \multicolumn{2}{c}{union} & \multicolumn{5}{c}{negation} \\
\cmidrule(lr){2-10} \cmidrule(lr){11-12} \cmidrule(lr){13-17}
\textbf{Dataset} & 1p & 2p & 3p & 4p & 2i & 3i & 4i & 1p2i & 2i1p & 2u & 2u1p & 2in & 3in & 2in1p & 2pi1pn & 2nu1p \\
\midrule
\multicolumn{17}{@{}l}{\textit{FB15k237, the fine-tuning dataset}} \\
\midrule
Training & $149{,}689$ & $149{,}689$ & $149{,}689$ & \hfill--\hfill & $149{,}689$ & $149{,}689$ & \hfill--\hfill & \hfill--\hfill & \hfill--\hfill & \hfill--\hfill & \hfill--\hfill & $14{,}968$ & $14{,}968$ & $14{,}968$ & $14{,}968$ & $14{,}968$ \\
Validation & $20{,}094$ & $5{,}000$ & $5{,}000$ & \hfill--\hfill & $5{,}000$ & $5{,}000$ & \hfill--\hfill & $5{,}000$ & $5{,}000$ & $5{,}000$ & $5{,}000$ & $5{,}000$ & $5{,}000$ & $5{,}000$ & $5{,}000$ & $5{,}000$ \\
\midrule
\multicolumn{17}{@{}l}{\textit{\famone}} \\
\midrule
FB15k237+H & $22{,}804$ & $936$ & $1{,}461$ & $2{,}670$ & $6{,}692$ & $9{,}615$ & $20{,}189$ & $1{,}586$ & $604$ & $6{,}695$ & $473$ & $5{,}000$ & $3{,}057$ & $2{,}663$ & $2{,}093$ & $5{,}000$ \\
NELL995+H & $17{,}021$ & $1{,}054$ & $1{,}583$ & $2{,}322$ & $8{,}278$ & $13{,}118$ & $22{,}326$ & $1{,}957$ & $1{,}260$ & $8{,}168$ & $1{,}073$ & $4{,}000$ & $5{,}236$ & $2{,}708$ & $1{,}995$ & $4{,}000$ \\
ICEWS18+H & $30{,}605$ & $2{,}125$ & $3{,}582$ & $4{,}338$ & $11{,}918$ & $27{,}252$ & $38{,}767$ & $14{,}746$ & $5{,}140$ & $7{,}516$ & $3{,}855$ & $1{,}548$ & $7{,}811$ & $2{,}875$ & $2{,}919$ & $1{,}508$ \\
\midrule
\multicolumn{17}{@{}l}{\textit{\famtwo}} \\
\midrule
FB 106\%+H & $5{,}446$ & $5{,}960$ & $2{,}340$ & $196$ & $8{,}227$ & $10{,}102$ & $9{,}086$ & $23{,}013$ & $3{,}481$ & $5{,}942$ & $1{,}273$ & $5{,}446$ & $1{,}119$ & $2{,}833$ & $1{,}037$ & $5{,}446$ \\
FB 113\%+H & $9{,}782$ & $3{,}607$ & $2{,}354$ & $142$ & $13{,}156$ & $23{,}277$ & $18{,}671$ & $23{,}753$ & $5{,}646$ & $9{,}361$ & $2{,}310$ & $6{,}379$ & $2{,}272$ & $1{,}232$ & $1{,}301$ & $6{,}379$ \\
FB 122\%+H & $14{,}458$ & $2{,}378$ & $5{,}810$ & $854$ & $9{,}498$ & $16{,}408$ & $24{,}835$ & $18{,}185$ & $7{,}424$ & $7{,}942$ & $4{,}794$ & $4{,}978$ & $3{,}931$ & $884$ & $970$ & $4{,}976$ \\
FB 134\%+H & $15{,}270$ & $3{,}332$ & $6{,}327$ & $444$ & $10{,}021$ & $16{,}790$ & $28{,}876$ & $22{,}155$ & $6{,}620$ & $7{,}874$ & $5{,}859$ & $5{,}343$ & $4{,}124$ & $1{,}397$ & $1{,}202$ & $5{,}343$ \\
FB 150\%+H & $17{,}683$ & $2{,}392$ & $6{,}170$ & $1{,}271$ & $9{,}066$ & $17{,}718$ & $28{,}279$ & $17{,}012$ & $7{,}570$ & $7{,}240$ & $4{,}878$ & $4{,}547$ & $5{,}029$ & $1{,}284$ & $1{,}025$ & $4{,}547$ \\
FB 175\%+H & $17{,}476$ & $3{,}350$ & $7{,}325$ & $2{,}180$ & $9{,}351$ & $18{,}914$ & $31{,}746$ & $22{,}411$ & $7{,}507$ & $7{,}289$ & $5{,}474$ & $5{,}124$ & $4{,}749$ & $1{,}188$ & $1{,}184$ & $5{,}125$ \\
FB 217\%+H & $16{,}902$ & $3{,}838$ & $7{,}556$ & $1{,}903$ & $9{,}932$ & $18{,}229$ & $28{,}265$ & $16{,}581$ & $8{,}002$ & $7{,}317$ & $6{,}643$ & $4{,}844$ & $4{,}760$ & $1{,}539$ & $1{,}616$ & $4{,}844$ \\
FB 300\%+H & $17{,}105$ & $2{,}190$ & $6{,}483$ & $2{,}263$ & $7{,}944$ & $15{,}667$ & $26{,}630$ & $15{,}841$ & $8{,}248$ & $6{,}218$ & $5{,}116$ & $4{,}541$ & $4{,}839$ & $1{,}057$ & $1{,}258$ & $4{,}545$ \\
FB 550\%+H & $13{,}670$ & $3{,}187$ & $7{,}274$ & $3{,}083$ & $11{,}569$ & $21{,}941$ & $33{,}753$ & $22{,}496$ & $8{,}691$ & $8{,}319$ & $6{,}540$ & $5{,}242$ & $4{,}165$ & $1{,}441$ & $1{,}656$ & $5{,}243$ \\
\midrule
\multicolumn{17}{@{}l}{\textit{\famthree}} \\
\midrule
WikiTopics Art+H & $4{,}638$ & $4{,}656$ & $4{,}879$ & $1{,}369$ & $5{,}487$ & $3{,}647$ & $1{,}614$ & $8{,}267$ & $2{,}529$ & $3{,}095$ & $1{,}085$ & $4{,}632$ & $906$ & $3{,}311$ & $3{,}754$ & $4{,}596$ \\
WikiTopics Health+H & $2{,}039$ & $3{,}747$ & $1{,}112$ & $260$ & $4{,}879$ & $5{,}198$ & $3{,}127$ & $5{,}817$ & $6{,}518$ & $3{,}080$ & $1{,}570$ & $2{,}039$ & $714$ & $4{,}457$ & $2{,}258$ & $2{,}025$ \\
WikiTopics Infrastructure+H & $3{,}756$ & $4{,}204$ & $458$ & $58$ & $2{,}401$ & $2{,}085$ & $916$ & $2{,}289$ & $3{,}791$ & $1{,}366$ & $1{,}077$ & $2{,}641$ & $296$ & $4{,}259$ & $4{,}141$ & $2{,}062$ \\
WikiTopics Location+H & $11{,}832$ & $4{,}248$ & $94$ & $15$ & $216$ & $89$ & $15$ & $325$ & $401$ & $106$ & $453$ & $6{,}322$ & $62$ & $340$ & $53$ & $5{,}615$ \\
WikiTopics People+H & $10{,}399$ & $6{,}750$ & $11{,}616$ & $231$ & $13{,}960$ & $9{,}352$ & $4{,}390$ & $13{,}591$ & $15{,}196$ & $7{,}356$ & $5{,}883$ & $8{,}127$ & $1{,}641$ & $3{,}015$ & $2{,}823$ & $7{,}995$ \\
WikiTopics Science+H & $1{,}759$ & $1{,}609$ & $720$ & $299$ & $2{,}421$ & $2{,}394$ & $1{,}867$ & $2{,}356$ & $1{,}341$ & $1{,}604$ & $764$ & $1{,}757$ & $529$ & $2{,}777$ & $2{,}778$ & $1{,}721$ \\
WikiTopics Taxonomy+H & $2{,}783$ & $1{,}113$ & $115$ & $26$ & $954$ & $1{,}132$ & $825$ & $512$ & $3{,}479$ & $630$ & $1{,}329$ & $1{,}077$ & $263$ & $283$ & $1{,}287$ & $944$ \\
\bottomrule
\end{tabular}
}
\end{table}

\end{document}